\documentclass[sigconf]{acmart}

\usepackage{booktabs} 
\copyrightyear{2019}
\acmYear{2019}
\setcopyright{iw3c2w3}
\acmConference[WWW '19]{Proceedings of the 2019 World Wide Web Conference}{May 13--17,
2019}{San Francisco, CA, USA}
\acmBooktitle{Proceedings of the 2019 World Wide Web Conference (WWW '19), May 13--17, 2019,
San Francisco, CA, USA}
\acmPrice{}
\acmDOI{10.1145/3308558.3313577}
\acmISBN{978-1-4503-6674-8/19/05}

\usepackage{helvet}  
\usepackage{url}  
\usepackage{balance}
\usepackage{makecell}
\usepackage{graphicx}
\usepackage{caption}
\usepackage{courier}
\usepackage{multirow}
\usepackage{color}
\usepackage{subcaption}
\usepackage{color}
\usepackage{todonotes}
\usepackage{amssymb}
\usepackage{gensymb}
\usepackage{enumitem}
\usepackage{multirow}
\usepackage[linesnumbered, ruled]{algorithm2e}
\usepackage{amsmath}

\newcommand{\argmax}{\arg\max}

\newcommand{\nop}[1]{}

\begin{document}
\title{Learning from Multiple Cities: A Meta-Learning Approach for \\Spatial-Temporal Prediction}
\author{Huaxiu Yao}
\affiliation{%
  \institution{Pennsylvania State University}}
\email{huaxiuyao@psu.edu}

\author{Yiding Liu}
\affiliation{%
  \institution{Nanyang Technological University}}
\email{liuy0130@e.ntu.edu.sg}

\author{Ying Wei}
\affiliation{%
  \institution{Tencent AI Lab}}
\email{judyweiying@gmail.com}

\author{Xianfeng Tang}
\affiliation{%
  \institution{Pennsylvania State University}}
\email{tangxianfeng@outlook.com}

\author{Zhenhui Li}
\affiliation{%
  \institution{Pennsylvania State University}}
\email{jessieli@ist.psu.edu}

\renewcommand{\shortauthors}{H. Yao et al.}

\begin{abstract}
Spatial-temporal prediction is a fundamental problem for constructing smart city, which is useful for tasks such as traffic control, taxi dispatching, and environment policy making. Due to data collection mechanism, it is common to see data collection with unbalanced spatial distributions. For example, some cities may release taxi data for multiple years while others only release a few days of data; some regions may have constant water quality data monitored by sensors whereas some regions only have a small collection of water samples. In this paper, we tackle the problem of spatial-temporal prediction for the cities with only a short period of data collection. We aim to utilize the long-period data from other cities via transfer learning. Different from previous studies that transfer knowledge from one single source city to a target city, we are the first to leverage information from multiple cities to increase the stability of transfer. Specifically, our proposed model is designed as a spatial-temporal network with a meta-learning paradigm. The meta-learning paradigm learns a well-generalized initialization of the spatial-temporal network, which can be effectively adapted to target cities. In addition, a pattern-based spatial-temporal memory is designed to distill long-term temporal information (i.e., periodicity). We conduct extensive experiments on two tasks: traffic (taxi and bike) prediction and water quality prediction. The experiments demonstrate the effectiveness of our proposed model over several competitive baseline models. 
\end{abstract}

\maketitle

\section{Introduction}
Recently, the construction of smart cities significantly changes urban management and services~\cite{wang2017region,zhao2017incorporating,zhao2017modeling,wei2018intellilight,liu2017experimental,liu2018efficient}. One of the most fundamental techniques in building smart city is accurate spatial-temporal prediction. For example, a traffic prediction system can help the city pre-allocate transportation resources and control traffic signal intelligently. An accurate environment prediction system can help the government develop environment policy and then improve the public's health.

Traditionally, basic time series models (e.g., ARIMA, Kalman Filtering and their variants)~\cite{shekhar2008adaptive,moreira2013predicting,lippi2013short}, regression models with spatial-temporal regularizations~\cite{ide2011trajectory,zheng2013time} and external context features~\cite{yi2018deep,wu2016interpreting} are used for spatial-temporal prediction. Recently, advanced machine learning methods (e.g., deep neural network based models) significantly improve the performance of spatial-temporal prediction~\cite{yao2018deep,yi2018deep,zhang2016deep} by characterizing non-linear spatial-temporal correlations more accurately. 
Unfortunately, the superior performance of these models is conditioned on 
large-scale training data which are probably inaccessible in real-world applications.
For example, there may exist only a few days of GPS traces for traffic prediction in some cities. 

Transfer learning has been studied as an effective solution to address the data insufficiency problem, by leveraging knowledge from those cities with abundant data (e.g., GPS traces covering a few years).
In~\cite{wei2016transfer}, the authors proposed to transfer semantically-related dictionaries learned from a data-rich city, i.e., a source city, to predict the air quality category in a data-insufficient city, i.e., a target city. The method proposed in~\cite{wang2018crowd} 
aligns similar regions across source and target cities for 
finer-grained transfer. However, these methods, 
transferring the knowledge from only a 
single source city, would cause unstable results and the risk of negative transfer. 
If the underlying data distributions are significantly different between cities, the knowledge transfer will make no contribution 
or even hurt the performance.

To reduce the risk, in this paper, we 
focus on transferring 
knowledge from \textit{multiple source cities} for the spatial-temporal prediction in a target city. Compared with single city, the knowledge extracted from multiple cities 
covers more 
comprehensive spatial-temporal correlations 
of a city, e.g., temporal dependency, spatial closeness, and region functionality, and thus increases the stability of transfer. 
However, this problem faces two key challenges. 
\begin{itemize}[leftmargin=*]\setlength{\itemsep}{0pt}

\item \textbf{How to adapt the knowledge to meet various scenarios of spatial-temporal correlations in a target city?} The spatial-temporal correlations in the limited data of a target city may vary considerably from city to city and even from time to time. For example, the knowledge to be transferred to New York is expected to differ from that to Boston. In addition, the knowledge to be transferred within the same city also differs from weekdays to weekends. Thus a sufficiently flexible algorithm capable of adapting the knowledge learned from multiple source cities to various scenarios is required.   

\item \textbf{How to capture and borrow long-period spatial-temporal patterns from source cities?} It is difficult to capture long-term spatial-temporal patterns (e.g., periodicity) accurately in a target city with limited data due to the effects of special events (e.g. holiday) or missing values. These patterns as indicators of region functionality, however, are crucial for 
spatial-temporal prediction~\cite{yao2018modeling,zonoozi2018periodic}. 
Take the traffic demand prediction as an instance. The traffic demand in residential areas is periodically high 
in the morning when people transit to work. Thus, it is promising but challenging to transfer such long-term patterns from source cities to a target city.
\end{itemize}

To address the 
challenges, 
we propose a novel 
framework for spatial-temporal prediction, namely \textbf{MetaST}. 
It is the first to incorporate the meta-learning paradigm into spatial-temporal networks (ST-net). 
The ST-net consists of a local CNN and an LSTM 
which jointly capture 
spatial-temporal features and correlations. 
With the meta-learning paradigm, we solve the first challenge by learning a well-generalized initialization of the ST-net 
from a large number of prediction tasks sampled from multiple source cities, which covers comprehensive spatial-temporal scenarios. 
Subsequently, the initialization can easily be adapted 
to a target city via fine-tuning, 
even when only 
few training 
samples are accessible. 
Second, we learn 
a global pattern-based spatial-temporal memory from all source cities, and transfer it to a target city to support long-term patterns.
The memory, describing and storing long-term spatial-temporal patterns, is jointly trained with the ST-net in an end-to-end manner. 

We evaluate the proposed framework 
on several datasets including taxi volume, bike volume, and water quality. The results show that our proposed MetaST consistently outperforms several baseline models. We 
summarize our contributions as follows. 
\begin{itemize}[leftmargin=*]
    \item To the best of our knowledge, we are the first to study the problem of transferring knowledge from multiple cities for the spatial-temporal prediction in a target city.
    \item We propose a novel MetaST framework to solve the problem 
    by combining a spatial-temporal network with the meta-learning paradigm.
    Moreover, we learn from all source cities 
    a global memory encrypting long-term spatial-temporal patterns, and 
    transfer it to further improve the spatial-temporal prediction in a target city. 
	\item We empirically demonstrate 
	the effectiveness of our proposed MetaST framework on three real-world spatial-temporal datasets.
\end{itemize}

The rest of this paper is organized as follows. We first review and discuss the related work in Section 2. Then, we define some notations and formulate the problem in Section 3. After that, we introduce details of the framework of MetaST in Section 4. We apply our model on three real-world datasets from two different domains and conduct extensive experiments in Section 5. Finally, we conclude our paper in Section 6.

\section{Related Work}
\label{sec:relatedwork}
In this section, we briefly review the works in two categories: some representative works for spatial-temporal prediction and knowledge transferring.
\subsection{Spatial-Temporal Prediction}
The earliest spatial-temporal prediction models are based on basic time series models (e.g., ARIMA, Kalman Filtering)~\cite{shekhar2008adaptive,moreira2013predicting,lippi2013short}. 
Recent studies further utilize external context data (e.g., weather condition, venue types, holiday, event information)~\cite{yi2018deep,wu2016interpreting,wang2017non} to enhance the prediction accuracy. Also, spatial correlation is taken into consideration by designing regularization of spatial smoothness~\cite{tong2017sim,xu2016gspartan,zheng2013time}. 

Recently, various deep learning methods have been used to capture complex non-linear spatial-temporal correlations and predict spatial-temporal series, such as stacked fully connected network~\cite{wang2017deepsd,yi2018deep}, convolutional neural network (CNN)~\cite{zhang2016deep,wang2017deep} and recurrent neural network (RNN)~\cite{yu2017deep}. Several hybrid models have been proposed to model both spatial and temporal information~\cite{yao2018deep,ke2017short,yao2018modeling}. These methods combine CNN and RNN, and achieve the state-of-the-art performance on spatial-temporal prediction. In addition, based on the road network structure, another type of hybird models combines graph aggregation mechanism and RNN for spatial-temporal prediction~\cite{zhang2018gaan,li2017graph,yu2017spatio} 

\emph{Different from previous studies of deep spatial-temporal prediction which all rely on a large set of training samples, we aim to transfer learned knowledge from source cities to improve the performance of spatial-temporal prediction in a target city with limited data samples.}

\subsection{Knowledge Transfer and Reuse}
Transfer learning and its related fields utilize previously learned knowledge to facilitate learning in new tasks when labeled data is scarce~\cite{naik1992meta,pan2010survey}. Previous transfer learning methods transfer different information from a source domain to a target domain, such as parameters~\cite{yang2007adapting}, instances~\cite{dai2009eigentransfer}, manifold structures~\cite{gong2012geodesic,gopalan2011domain}, deep hidden feature representations~\cite{tzeng2017adversarial,yosinski2014transferable}. Recently, meta-learning (a.k.a., learning to learn) transfers shared knowledge from multiple training tasks to a new task for quick adaptation. These techniques include learning a widely generalizable initialization~\cite{finn2017model,liu2018transductive}, optimization trace~\cite{ravi2016optimization}, metric space~\cite{snell2017prototypical}, transfer learning skills~\cite{ying2018transfer}. 

However, only a few attempts have been made on transferring knowledge on the space. \cite{wei2016transfer} proposes a multi-modal transfer learning framework for predicting air quality category, which combines multi-modal data by learning a semantically coupled dictionary for multiple modalities in a source city. \emph{However, this method works on multimodal features instead of spatial-temporal sequences we focus on. Therefore, it cannot be applied to solve the problem.} For predicting traffic flow, \cite{wang2018crowd} leverages the similarity of check-in records/spatial-temporal sequences between a source city and a target city to construct the similarity regularization for knowledge transfer. \emph{Different from this method that intelligently learns the correlation which could be linear or non-linear. Compared with both methods above, in addition, our model transfers the shared knowledge from multiple cities to a new city, which increase the stability of transfer and prediction.}

\section{Definitions and Problem Formulation}
In this section, we define some notations used in this paper and formally define the setup of our problem. 

\noindent\textbf{Definition 1 (Region)} Following 
previous works~\cite{zhang2016deep,yao2018deep}, we spatially divide a 
city $c$ 
into an $I_{c}\times J_{c}$ 
grid map which contains $I_{c}$ rows and $J_{c}$ columns. 
We treat each grid as a region $r_{c}$, and define the set of all regions as $\mathcal{R}_c$. 

\noindent\textbf{Definition 2 (Spatial-Temporal Series)}
In city $c$, 
we denote the current/latest timestamp as $k_c$, and the time range as a set 
$\mathcal{K}_{c}=\{k_c-|\mathcal{K}_c|+1,..., k_c\}$
consisting of $\vert\mathcal{K}_c\vert$ evenly split non-overlapping time intervals. 
Then, the spatial-temporal series in city $c$ is represented as 
\begin{equation}
\mathcal{Y}_c=\{y_{r_c,k_c}|r_c\in \mathcal{R}_c, k_c\in \mathcal{K}_c\},
\end{equation}
where $y_{r_c, k_c}$  is 
the spatial-temporal information to be predicted (e.g., traffic demand, air quality, climate value). 

\noindent\textbf{Problem Definition} 
Suppose that we have a set of source cities $\mathcal{C}_s=\{c_1,..., c_S\}$ 
and a target city $c_t$
with insufficient data 
(i.e., $\forall s\!\in\!\{1,\!\cdots\!,S\}$, $\vert\mathcal{K}_{c_s}\vert\!\gg\! \vert\mathcal{K}_{c_t}\vert$), our goal is to predict the spatial-temporal information of the next timestamp $k_{c_t}+1$ in the testing dataset of the target city $c_t$, i.e.,  
\begin{equation}
y_{r_{c_t}, k_{c_t}+1}^*={\argmax}_{y_{r_{c_t}, k_{c_t}+1}} p(y_{r_{c_t}, k_{c_t}+1}|\mathcal{Y}_{c_t}, f_{\theta_{0}}),
\end{equation}
where $f$ represents the ST-net which serves as the base model to predict the spatial-temporal series. Detailed discussion about ST-net is in given Section~\ref{sec:localst}. More importantly, in the meta-learning paradigm, $\theta_{0}$ denotes the initialization for all parameters of the ST-net, which is adapted from $\mathcal{Y}_{c_1},\cdots,\mathcal{Y}_{c_S}$. 

\section{Methodology}
In this section, we elaborate our proposed method \textbf{MetaST}. In particular, we first introduce the ST-net as the base model 
$f$ for spatial-temporal prediction, and then present our proposed spatial-temporal meta-learning framework for knowledge transfer. 

\subsection{Spatial-Temporal Network}
\label{sec:localst}
Recently, hybrid models 
combining convolution neural networks (CNN)~\cite{krizhevsky2012imagenet} and LSTM~\cite{hochreiter1997long} have achieved the state-of-the-art performances on spatial-temporal prediction. 
Thus, following~\cite{yao2018deep}, we adopt a CNN to capture the spatial dependency between regions, and an LSTM to model the temporal evolvement of each region.

In city $c$, for each region $r_c$ at time $k_c$, we regard the spatial-temporal value of this region and its surrounding neighbors as an $N\times N$ image with $v$ channels $\mathbf{Y}_{r_c, k_c} \in \mathbb{R}^{N\times N\times v}$, where 
region $r_c$ is in the center of this image. For instance, when N=3, we are considering a center cell as well as 8 adjacent grid cells of the cell, which is a 3*3 size neighborhood. The number of 
channels $v$ depends on a 
specific task. For example, in taxi volume prediction, we jointly predict taxi pick-up volume and drop-off volume, so that 
the number of 
channels 
equals two, i.e., $v=2$. 
Taking 
$\mathbf{Y}_{r_c, k_c}$ as input $\mathbf{Y}^0_{r_c, k_c}$, 
a local CNN  
computes the $q$-th layer progressively: 
\begin{equation}
\label{eq:spatialcnn}
\mathbf{Y}_{r_c, k_c}^{q}=\mathrm{ReLU}(\mathbf{W}_{r}^{q} \ast\mathbf{Y}_{r_c, k_c}^{q-1} +\mathbf{b}_{r}^{q}),
\end{equation}
where $\ast$ represents the convolution operation, $\mathbf{W}_{r}^{q}$ and $\mathbf{b}_{r}^{q}$ are learnable parameters. After $Q$ convolutional layers, 
a fully connected layer following a flatten layer is used to infer the spatial representation of region $r_c$ as 
$\mathbf{s}_{r_c, k_c}$ eventually.

\begin{figure}[h]
	\centering
	\includegraphics[width=0.46\textwidth]{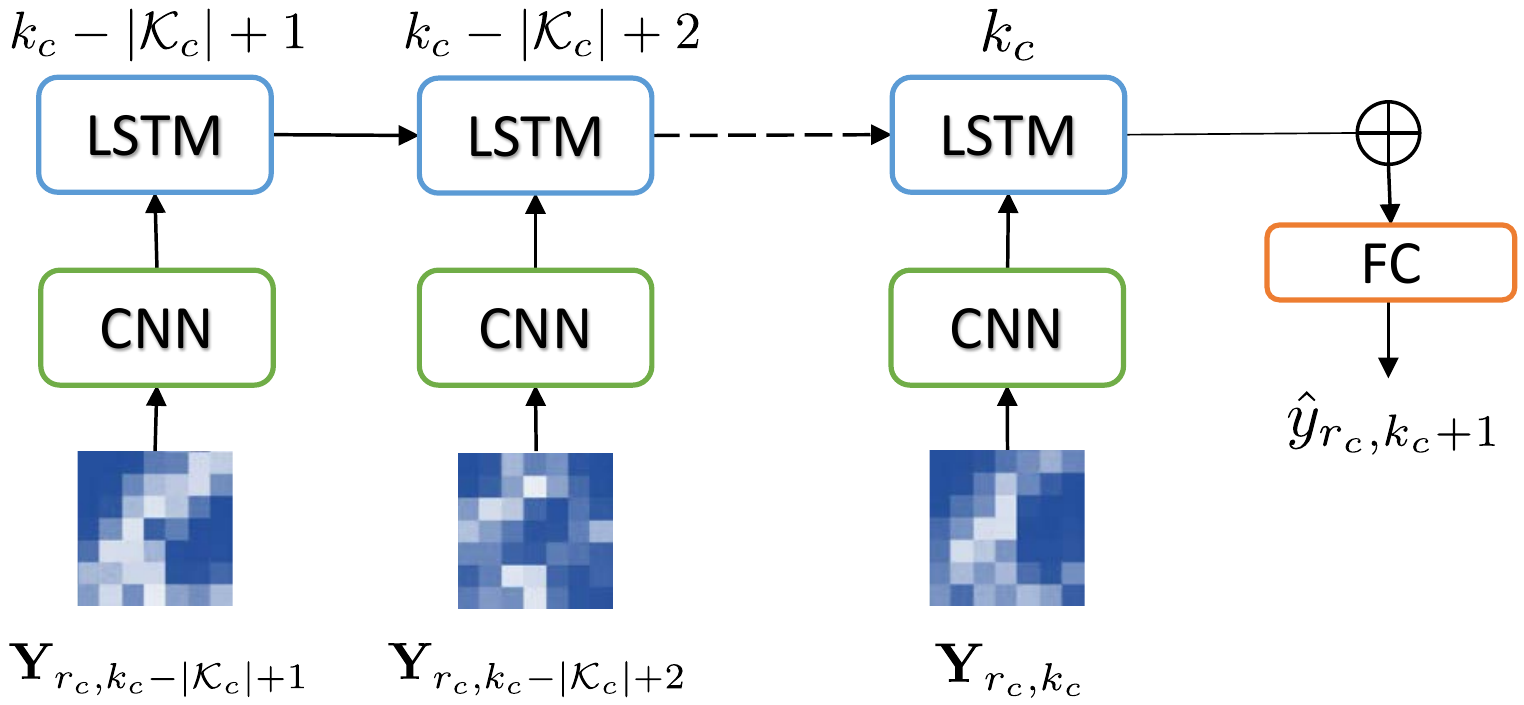}
	\caption{The framework of the ST-net. The heatmaps are exemplar spatial-temporal values. CNN is used to capture spatial dependency and then LSTM is used to handle temporal correlation.}
	\label{fig:stnet}
\end{figure}

Then, 
for predicting $y_{r_c, k_c+1}$, 
we model the temporal unfolding of region $r_c$ by passing all the 
spatial representations 
along the time span $\{k_c-|\mathcal{K}_c|+1,..., k_c\}$ through an LSTM, which can be formulated as

\begin{equation}
\begin{split}
&{\bf i}_{r_c, k_c} = \sigma ({\bf U}_{i}[\mathbf{s}_{r_c, k_c}; \mathbf{e}_{r_c, k_c}] + {\bf W}_{i}{\bf h}_{r_c, k_c-1} + {\bf b}_{i}), \\
&{\bf f}_{r_c, k_c} = \sigma ({\bf U}_{f}[\mathbf{s}_{r_c, k_c}; \mathbf{e}_{r_c, k_c}] + {\bf W}_{f}{\bf h}_{r_c, k_c-1} + {\bf b}_{f}), \\
&{\bf d}_{r_c, k_c} = \sigma ({\bf U}_{d}[\mathbf{s}_{r_c, k_c}; \mathbf{e}_{r_c, k_c}] + {\bf W}_{d}{\bf h}_{r_c, k_c-1} + {\bf b}_{d}), \\
&{\bf \hat{c}}_{r_c, k_c} = \tanh({\bf U}_{c}[\mathbf{s}_{r_c, k_c}; \mathbf{e}_{r_c, k_c}] + {\bf W}_{c}{\bf h}_{r_c, k_c-1}+ {\bf b}_{c}), \\
&{\bf c}_{r_c, k_c} = {\bf f}_{r_c, k_c}\circ{\bf c}_{r_c, k_c-1} + {\bf i}_{r_c, k_c}\circ {\bf \hat{c}}_{r_c, k_c}, \\
& {\bf h}_{r_c, k_c} = \tanh({\bf c}_{r_c, k_c})\circ{\bf d}_{r_c, k_c},
\end{split}
\label{eq:lstm}
\end{equation}
where $\circ$ denotes Hadamard product, ${\bf U}_{j}$, ${\bf W}_{j}$, and ${\bf b}_{j}$ ($j \in \{i, f, d, c\}$) are learnable parameters, ${\bf i}_{r_c, k_c}$, ${\bf f}_{r_c, k_c}$, and ${\bf d}_{r_c, k_c}$ are forget gate vector, input gate vector, and output gate vector of the $i$-th context feature, respectively. $\mathbf{h}_{r_c, k_c}$ denotes the spatial-temporal representation of region $r_c$, and $|\mathcal{K}_c|$ is the number of time steps we use to consider the temporal information. Note that $\mathbf{e}_{r_c, k_c}$ represents other external features (e.g., weather, holiday) that can be incorporated, if applicable. 
By doing these, $\mathbf{h}_{r_c, k_c}$ encodes both the spatial and temporal information of region $r_c$.
As a result, 
the value of the next time step of spatial-temporal series, i.e., $\hat{y}_{r_c, k_c+1}$,  can be predicted by 
\begin{equation}
\label{eq:finalfc}
	\hat{y}_{r_c, k_c+1}=\tanh(\mathbf{W}_n\mathbf{h}_{r_c, k_c}+\mathbf{b}_n),
\end{equation}
where $\mathbf{W}_n$ and $\mathbf{b}_n$ are learnable parameters. The output is scaled to (-1,1) via a $\tanh(\cdot)$ function, 
to be consistent with the normalized spatial-temporal values. 
We later denormalize the prediction to get the
actual demand values.
We formulate the loss function of ST-net for each city $c$ as:
\begin{equation}
\label{eq:lossmse}
\mathcal{L}_{c}=\sum_{r_c}\sum_{k_c} (\hat{y}_{r_c, k_c+1} - y_{r_c, k_c+1})^2,
\end{equation}
so as to enforce the estimated spatial-temporal value to be as close to the ground-truth $y_{r_c, k_c+1}$ as possible. 
As introduced previously, 
we denote all the parameters of the ST-net as $\theta$, and the ST-net parameterized by $\theta$ as $f_\theta$. For better illustration, we visualize the structure of the spatial-temporal network (\textbf{ST-net}) in Figure~\ref{fig:stnet}.
\subsection{Knowledge Transfer}
\begin{figure*}[!t]
	\centering
	\includegraphics[width=0.85\textwidth]{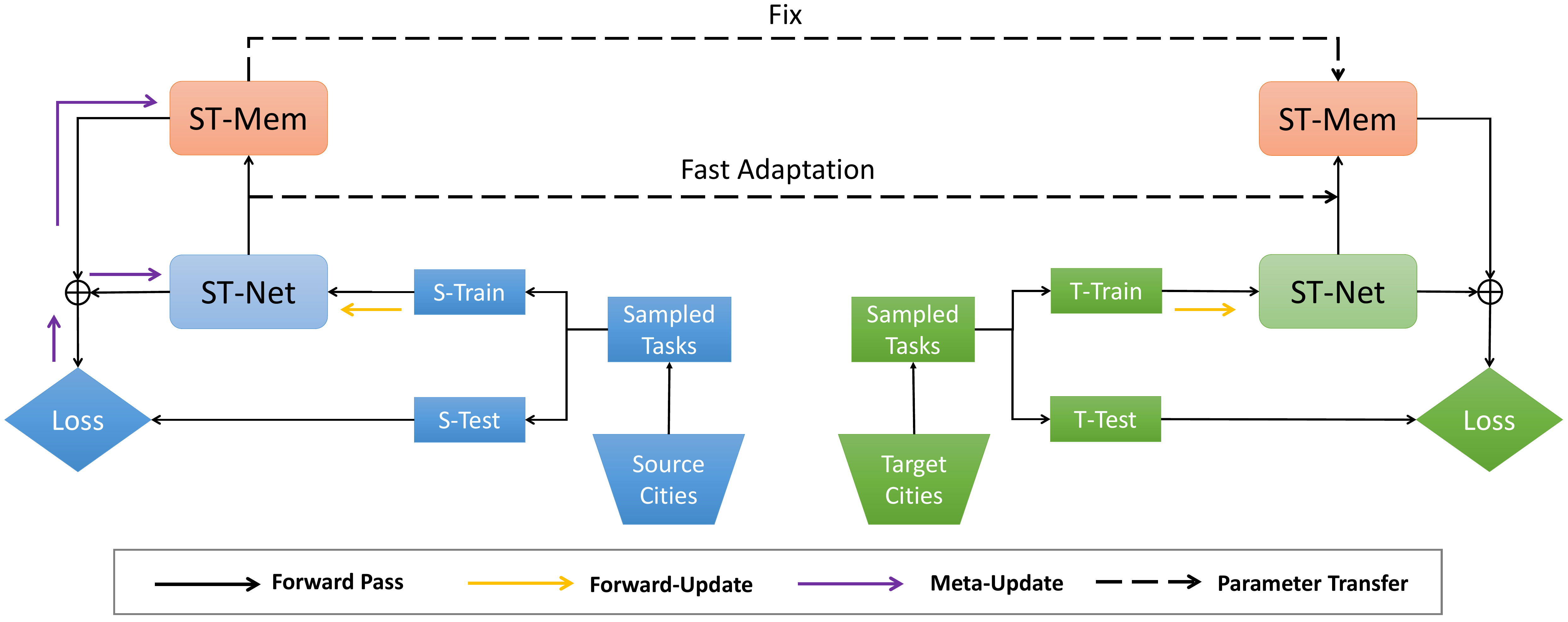}
	\caption{The framework of proposed MetaST. ST-net and ST-mem mean spatial-temporal network and spatial-temporal memory. S-train and S-test, T-train and T-test represent the training and testing set of source tasks (i.e., source cities) and target tasks (i.e., source cities), respectively. In the process of knowledge transfer, the parameters of ST-net will be updated by the training set in target city (i.e., T-train), while the parameters of ST-mem are fixed.}
	\label{fig:framework}
\end{figure*}

Next, we propose a meta-learning framework that enables our ST-net model to borrow 
knowledge from multiple cities. The framework consists of two parts: adapting the initialization 
and learning the spatial-temporal memory. We present the whole framework 
in Figure~\ref{fig:framework}.
\subsubsection{Adapting the Initialization}
As we described before, we are supposed to increase the stability of prediction by transferring knowledge from multiple source cities. Since the spatial-temporal correlation of limited data in a target city may vary considerably from city to city and even from time to time. For example, in traffic prediction, the knowledge to be transferred to Boston with limited weekend data is expected to differ from that to Chicago with limited weekday data. Thus, the extracted knowledge from multiple cities is expected to include comprehensive spatial-temporal correlations such as spatial closeness and temporal dependency, so that we can adapt the knowledge to a target city with limited data under different scenarios.

In ST-net, the parameters $\theta$ are exactly the knowledge which encrypts spatial-temporal correlations. To effectively adapt the parameters to a target city, as suggested in model-agnostic meta-learning (MAML)~\cite{finn2017model}, 
initialization of $\theta$ from multiple source cities, i.e., $\theta_0$, so that the ST-net initialized by $\theta_0$ achieves the minimum of the average of generalization losses over all source cities, i.e., 
\begin{equation}
\label{eq:trainloss_pre}
\theta_0 = \min_{\theta_0}\sum_{c_s \in \mathcal{C}_s}\mathcal{L}^{'}_{c_s}(f_{\theta_0-\alpha \nabla_{\theta}\mathcal{L}_{c_s}(f_{\theta})}).
\end{equation}

Here we would note that 
$\mathcal{L}_{c_s}(f_{\theta_{c_s}})$ denotes the training loss on the training set of a city $c_s$ sampled from $\mathcal{C}_{s}$, i.e., $\mathcal{D}_{c_s}$ (refer to S-train in Figure~\ref{fig:framework}).
We illustrate the iterative update of the parameters $\theta_{c_s}$ during the training process (shown as the yellow solid arrow in Figure~\ref{fig:framework}), by showing one exemplar gradient descent, i.e.,
\begin{equation}
\label{eq:innertrain}
\theta_{c_s}=\theta_0-\alpha \nabla_{\theta}\mathcal{L}_{c_s}(f_{\theta}).
\end{equation}
In practice, we can use several steps of gradient descent to update from the initialization $\theta_0$ to $\theta_{c_s}$. For each city $c_s$, the training process is repeated on batches of series sampled from $D_{c_s}$.

Since Eq.~\eqref{eq:trainloss_pre} minimizes the generalization loss, $\mathcal{L}^{'}_{c_s}(\cdot)$ evaluates the loss on the test set of the city $c_s$, i.e., 
$\mathcal{D}^{'}_{c_s}$ (refer to S-test in Figure~\ref{fig:framework}).
By optimizing the problem in Eq.~(\ref{eq:trainloss_pre}) using stochastic gradient descent (shown as the purple solid arrow in Figure~\ref{fig:framework}), we obtain an initialization which can generalize well on different source cities. 
Therefore, it is widely expected that transferring the initialization $\theta_0$ to a target city $c_t$ would also bring superior generalization performance, which we will detail in the end of this section. 
\subsubsection{Spatial-Temporal Memory}
\begin{figure}[!t]
	\centering
 	\includegraphics[width=0.45\textwidth]{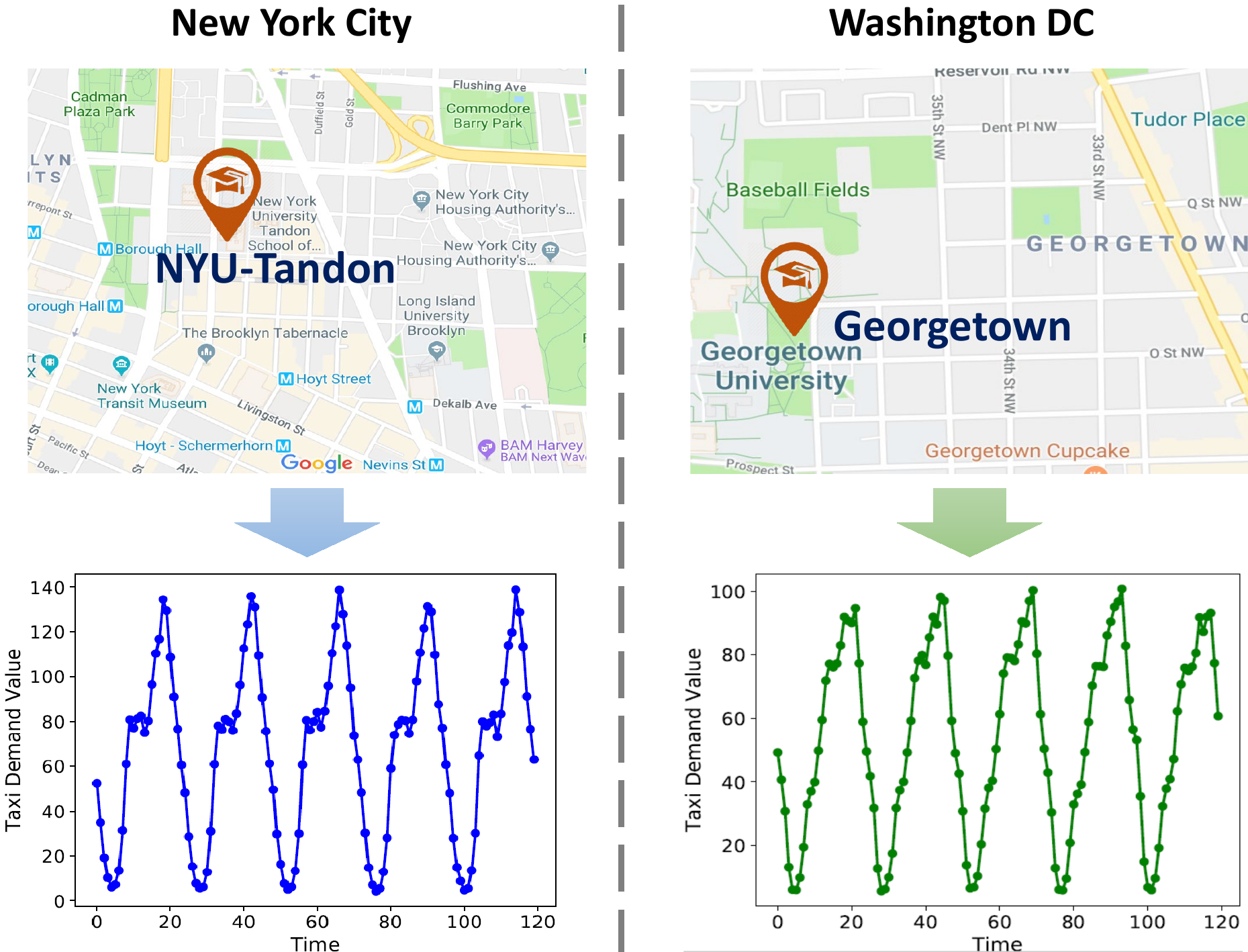}
	\caption{The illustration of spatial-temporal patterns of two regions and their corresponding region functionality.}
	\label{fig:func_region}
\end{figure}

\begin{figure}[!t]
	\centering
 	\includegraphics[width=0.48\textwidth]{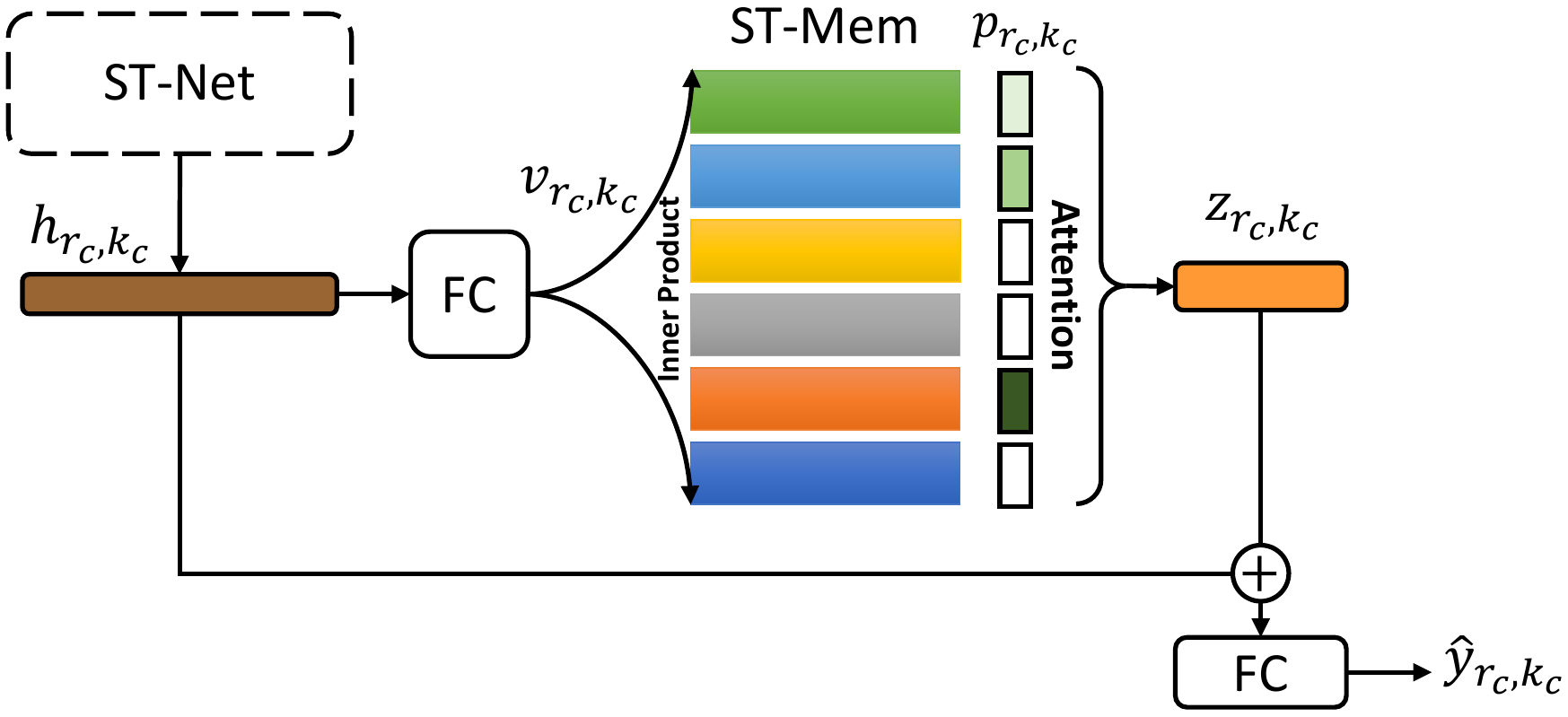}
	\caption{The stucture of ST-mem. The spatial-temporal representation $\mathbf{h}_{r_c,k_c}$ is projected as a query vector $\mathbf{v}_{r_c,k_c}$. Then, the attention mechanism is used to get the pattern representation $\mathbf{z}_{r_c,k_c}$. The short-term spatial-temporal representation and pattern representation are concatenated together for prediction.}
	\label{fig:mem_kd}
	\vspace{-1em}
\end{figure}

In spatial-temporal prediction, long-term spatial-temporal patterns (e.g., periodic patterns) play an important role~\cite{zonoozi2018periodic,yao2018modeling}. These patterns reflect the spatial functionality of each region and can be regarded as the global property shared by different cities. An example of spatial-temporal patterns and their corresponding regions are shown in Figure~\ref{fig:func_region}. 
In this figure, one region around NYU-Tardon in New York City and another one around Georgetown University in Washington DC are selected. The averaged 5-days' taxi demand distributions of both regions are daily periodic and similar, whose value are higher in the afternoon when students and faculties go back to home. The similarity of distributions between different cities verifies our assumption, i.e., spatial functionality is globally shared.
However, in a target city, these patterns are hard to be captured with limited data due to missing values or the effects of special events (e.g., holidays). 
Thus, we propose a global memory, named \textbf{ST-mem}, to store long-term spatial-temporal patterns and further transfer to improve prediction accuracy in target cities. The framework of ST-Mem is illustrated in Figure~\ref{fig:mem_kd}. 

Based on spatial-temporal patterns, we first cluster all the regions in source cities to $G$ categories. The categories $G$ of regions represent different region functionalities. For region $r_c$, the clustering results are denoted as $\mathbf{o}_{r_c} \in \mathbb{R}^G$. If region $r_c$ belongs to cluster $g$, $\mathbf{o}_{r_c}(g)=1$, otherwise $\mathbf{o}_{r_c}(g)=0$. Then, we construct a parameterized spatial-temporal memory $\mathcal{M}\in \mathbb{R}^{G\times d}$. Each row of the memory stores the pattern representation of a category, and the dimension of the pattern representation is $d$. 

Next, we utilize the knowledge of patterns stored in memory $\mathcal{M}$, and distill this knowledge for prediction via attention mechanism~\cite{luong2015effective}.
Since we only have short-term data in a target city, we use the representation of short-term data to query ST-mem. In particular, we linearly project the spatial-temporal representation $\mathbf{h}_{r_c, k_c}$ of ST-net to get the query vector for the attention mechanism, which is formulated as:
\begin{equation}
    \mathbf{v}_{r_c, k_c}=\mathbf{W}_l\mathbf{h}_{r_c, k_c}+\mathbf{b}_l.
\end{equation}
Then, we use the query vector $\mathbf{v}_{r_c, k_c}\in \mathbb{R}^{d}$ to calculate the similarity score between it and the pattern representation for each category $g$. Formally, the similarity score is defined as
\begin{equation}
    \mathbf{p}_{r_c, k_c}(g)=\frac{\exp(
\langle\mathbf{v}_{r_c,k_c},\mathcal{M}(g)\rangle)}{\sum_{g^{'}=1}^G\exp(\langle\mathbf{v}_{r_c,k_c},\mathcal{M}(g^{'})\rangle)},
\end{equation}
where $\mathcal{M}(g)$ means the $g$-th row of memory (i.e., for the $g$-th pattern category). We calculate the representation of spatial-temporal pattern $\mathbf{z}_{r_c, k_c}$ as:
\begin{equation}
\label{eq:attention}
    \mathbf{z}_{r_c, k_c} = \sum_{g=1}^G \mathbf{p}_{r_c, k_c}(g)*\mathcal{M}(g).
\end{equation}
Then, we concatenate the representation $\mathbf{z}_{r_c, k_c}$ of spatial-temporal patterns with the representation $\mathbf{h}_{r_c, k_c}$ of ST-net. And the input of the final layer $\mathbf{h}_{r_c, k_c}$ in Eq.~\eqref{eq:finalfc} is replaced by the enhanced representation i.e.,  
\begin{equation}
\label{eq:finalfc2}
	\hat{y}_{r_c, k_c+1}=\tanh(\mathbf{W}_n^{'}[\mathbf{h}_{r_c, k_c};\mathbf{z}_{r_c, k_c}]+\mathbf{b}^{'}_n).
\end{equation}

where $\mathbf{W}_n^{'}$ and $\mathbf{b}^{'}_n$ are learnable parameters. 

In order to learn the memory $\mathcal{M}$, we construct the clustering loss of city $c_s$, which enforce the attention scores to be consistent with previous clustering results $\mathbf{o}_{r_c}$. The formulation of the clustering loss is as follows:
\begin{equation}
\mathcal{L}_{clu}=-\sum_{r_c}\sum_{k_c}\mathbf{o}_{r_c} \log(\mathbf{p}_{r_c, k_c}).
\end{equation}
Additionally, the memory $\mathcal{M}$ is also updated when we train the MetaST framework, together with the initialization $\theta_0$. Thus, we revise the loss function in Eq.~\eqref{eq:trainloss_pre} by adding the clustering loss. Then, our final objective function is:
\begin{equation}
\label{eq:loss}
\theta_0, \mathcal{M}=\min_{\theta_0, \mathcal{M}} \sum_{c_s\in \mathcal{C}_{s}} \gamma \mathcal{L}_{clu} (f_{\theta_{c_s }})+\mathcal{L}^{'}_{c_s}(f_{\theta_{c_s}}),
\end{equation}
where $\gamma$ is a trade-off hypeparameter and is used to balance the effect of each part. $\theta_{c_s}=\theta_0-\alpha \nabla_{\theta}\mathcal{L}_{c_s}(f_{\theta})$ is defined in Eq.~\eqref{eq:innertrain}. By using testing set $\mathcal{D}^{'}_{c_s}$ of each city $c_s$, the memory $\mathcal{M}$ and initialization $\theta_0$ are updated by gradient descent.
Note that, as we discussed before, the spatial-temporal patterns are common property between cities. We do not update memory $\mathcal{M}$ when training a specific task (i.e., Eq.~\eqref{eq:innertrain}). In Figure~\ref{fig:framework}, the purple solid arrow indicates the process of meta-update. In Eq.~\eqref{eq:loss}, the initialization $\theta_0$ and memory $\mathcal{M}$ are mutually constrained. Since the memory $\mathcal{M}$ provides region-level relationship via spatial-temporal patterns, it can also help learn the initialization $\theta_0$ of ST-net. 
\subsubsection{Transfer Knowledge to Target Domain} 
To improve the prediction in target cities, we transfer the ST-mem $\mathcal{M}$ and initialization $\theta_0$ of ST-net. The black dotted line in Figure~\ref{fig:framework} shows the process of knowledge transfer. For each new city $c_t$ sampled from $\mathcal{C}_{t}$, the memory is fixed and the parameters $\theta_{c_t}$ are trained with initialization $\theta_0$ and training samples $\mathcal{D}_{c_t}$ (i.e., T-train in Figure~\ref{fig:framework}), which is defined as:
\begin{equation}
\label{eq:innertest}
\theta_{c_t}=\theta_0-\alpha \nabla_{\theta}\mathcal{L}_{c_t}(f_{\theta}),
\end{equation}
where $\mathcal{L}_{c_t}$ is the loss function of training set in target city $t$ and the formulation is:
\begin{equation}
    \mathcal{L}_{c_t}=\sum_{r_t}\sum_{k_t} (\hat{y}_{r_t, k_t+1} - y_{r_t, k_t+1})^2.
\end{equation}
The predicted value $\hat{y}_{r_t, k_t+1}$ is calculated via Eq.~\eqref{eq:finalfc2}. Hence, we distill knowledge of spatial-temporal patterns from source cities to target city via ST-mem $\mathcal{M}$. Finally, we evaluate the model $f_{\theta_{c_t}}$ on testing set $\mathcal{D}^{'}_{c_t}$ (i.e., T-test in Figure~\ref{fig:framework}) of city $c_t$ and get the prediction value of this city. The whole framework of MetaST is shown in Algorithm 1.

\begin{algorithm}[t]
	\caption{Framework of MetaST}
	\label{alg:outline}
	\KwIn{Set of target cities $\mathcal{C}_{t}$; Set of source cities $\mathcal{C}_{s}$; hyperparameter $\gamma$}
	\KwOut{Spatial-temporal predictions of each target city}
	Randomly initialize $\theta_0$ and $\mathcal{M}$\;
	Cluster all regions in $\mathcal{C}_{s}$ and get $\mathbf{o}_{r_c}$ for each region $r_c$\;
	\tcc{learning $\theta_0$ and $\mathcal{M}$ on source cities}
	\While{not done}{
		Sample a batch of cities from $\mathcal{C}_{s}$\;
		\For{each city $c_s$}{
			Sample a set $\mathcal{D}_{c_s}$ from $c_s$\;
			Evaluate $\nabla \mathcal{L}_{c_s}(f_{\theta_{c_s}})$ using $\mathcal{D}_{c_s}$ by Eq.~\eqref{eq:lossmse}\;
			Compute adapted parameters $\theta_{c_s}$ with gradient descent by Eq.~\eqref{eq:innertrain}\;
			Sample a new set $\mathcal{D}^{'}_{c_s}$ from $c_s$\;
			Evaluate $\mathcal{L}^{'}_{c_s}(f_{\theta_s})$, $\mathcal{L}_{clu}(f_{\theta_{c_s}})$ by $\mathcal{D}^{'}_{c_s}$\;
            }
		Update $\theta_0$, $\mathcal{M}$ by gradient descent\;

}
\tcc{Evaluate our model on target cities}
\For{each city $c_t$ in $\mathcal{C}_{t}$}{
		Sample a sample set $\mathcal{D}_{c_t}$ from $c_t$\;
		Compute adapted parameters $\theta_{c_t}$ with gradient descent by Eq.~\eqref{eq:innertest}\;
		Sample new series $\mathcal{D}^{'}_{c_t}$ and predict\;
	}
\label{alg:metast}
\end{algorithm}

\section{Experiment}
In this section, we conduct extensive experiments for two domain applications to answer the following research questions:

\begin{itemize}[leftmargin=0.15in]\setlength{\itemsep}{0pt}
\item Whether MetaST can outperform baseline methods for  inference tasks, i.e., traffic volume prediction in and water quality (pH value) prediction?
\item How do various hyper-parameters, e.g., the dimensions of each cluster in ST-mem or trade-off factor, impact the model's performance? 
\item Whether ST-mem can detect distinguished spatial-temporal patterns?
\end{itemize}

\subsection{Application-$\uppercase\expandafter{\romannumeral1}$: Traffic Prediction}
\subsubsection{Problem Overview} 
We first conduct experiments on two traffic prediction tasks, i.e., taxi volume prediction and bike volume prediction. Similar as the previous traffic prediction task in~\cite{yao2018modeling,zhang2016deep}, each individual trip departs from a region, and then arrives at the destination region. Thus, our task is to predict the pick-up (start) and drop-off (end) volume of taxi (and bike) at each time interval for each region. The time inveral of traffic prediction task is one hour. We use \emph{Root Mean Square Error} (RMSE) to evaluate our model, which is defined as:
\begin{equation}
\mathrm{RMSE}=\frac{1}{|N|}\sqrt{\sum_{r_t}\sum_{k_t} (\hat{y}_{r_t, k_t+1} - y_{r_t, k_t+1})_2^2},
\end{equation}
where $\hat{y}_{r_t, k_t+1}$ and $y_{r_t, k_t+1}$ represent predicted value and actual value, respectively. $|N|$ means the number of all samples in testing set.
\subsubsection{Data Description}
For taxi volume prediction, we collect five representative mobility datasets from five different cities to evaluate
the performance of our proposed model, i.e., New York City (NYC)\footnote{http://www.nyc.gov/html/tlc/html/about/trip\_record\_data.shtml}, Washington DC (DC), Chicago (CHI), Porto\footnote{https://www.kaggle.com/c/pkdd-15-predict-taxi-service-trajectory-i/data}, and Boston (BOS). We use NYC, DC, Porto as the source cities and CHI, BOS as the target cities. Note that the Boston data does not have drop-off records, and thus we only report the results on predicting pick-up volume.

For bike volume prediction, we collect three datasets from four cities, i.e., NYC\footnote{https://www.citibikenyc.com/system-data}, DC\footnote{https://www.capitalbikeshare.com/system-data}, and CHI\footnote{https://www.divvybikes.com/system-data}. NYC and DC are used as source cities, and CHI are used as target city. All trips in the above datasets contain time and geographic coordinates of pick-up and drop-off. For each city above, as discussed before, we spatially divide them to a grid map. The grid map size of NYC, DC, CHI, BOS, Porto is $10\times20$, $16\times 16$, $15\times 18$, $18\times15$, $20\times 10$, respectively. Detailed statistics of these datasets are listed in Table~\ref{tab:trafficdata}.  

In addition, for each source city, we select 80\% data for training and validation, and the rest for testing. For each target city, we select the 1-day, 3-day and 7-day data for training, and the rest for testing.
\begin{table}[t]
	\caption{Data Statistics of Traffic Prediction}
	\centering
	\begin{tabular}{|l|l|c|c|c|}
		\hline
		Data                      & City  & \multicolumn{1}{l|}{Time Span (m/d/y)} & \multicolumn{1}{l|}{Trips} & \multicolumn{1}{l|}{Size} \\ \hline
		\multirow{5}{*}{Taxi}         & NYC   & 1/1/15-7/1/15              & 6,748,857             & 10$\times$20     \\ \cline{2-5} 
		& DC    & 5/1/15-1/1/16              & 8,151,077            &   16$\times$16   \\ \cline{2-5} 
		& Porto &         7/1/13 - 6/30/14                       &       1,710,671          &    20$\times$10       \\ \cline{2-5} 
		& CHI   & 9/1/13-11/1/14             & 124,820             &    15$\times$18   \\ \cline{2-5} 
		& BOS   &            10/1/12 - 10/31/12                    &      839,897        & 18$\times$15              \\ \hline\hline
		\multirow{4}{*}{Bike} & NYC   & 1/1/17-12/31/17            &     16,364,475       & 10$\times$20                \\ \cline{2-5} 
		& DC    & 1/1/17-12/31/17            &            3,732,263     & 16$\times$16           \\ \cline{2-5} 
		& CHI   & 1/1/17-2/1/17            &         106,165     & 15$\times$18              \\ \hline
	\end{tabular}
	\label{tab:trafficdata}
	\vspace{-1em}
\end{table}

\subsubsection{Compared Baselines}
We compare our model with the following two categories of methods: non-transfer methods and transfer methods. Note that, for non-transfer baselines, we only use the training data of target cities (limited training data) to train the model and use the testing data to evaluate it. For transfer baselines, we transfer the learned knowledge from source cities to improve the prediction accuracy.~\\
\textbf{Non-transfer baselines:}
\begin{itemize}[leftmargin=*]\setlength{\itemsep}{0pt}
    \item \textbf{Historical Average (HA)}: For each region, HA predicts spatial-temporal value based on the average value of the previous relative time. For example, if we want to predict the pick-up volume at 9:00am-10:00am, HA is the average value of all time intervals from 9:00am to 10:00am in training data.
    \item \textbf{ARIMA}: Auto-Regressive Integrated Moving Average (ARIMA) is a traditional time-series prediction model, which considers the temporal relationship of data.
    \item \textbf{ST-net}: This method simply use the spatial-temporal neural network defined in Section~\ref{sec:localst} to predict traffic volume. Both pick-up and drop-off volume are predicted together. 
\end{itemize}
\textbf{Transfer baselines:}
\begin{itemize}[leftmargin=*]\setlength{\itemsep}{0pt}
    \item \textbf{Fine-tuning Methods}: We include two types of fine-tuning methods: (1) Single-source fine-tuning (\textbf{Single-FT}): train ST-net in one source city (e.g., NYC, DC or Porto in taxi data) and fine-tune the model for target cities; 
    and (2) multi-source fine-tune (\textbf{Multi-FT}): mix all samples from all source cities and fine-tune the model in target cities. 
    \item \textbf{RegionTrans~\cite{wang2018crowd}}: RegionTrans transfers knowledge from one city to another city for traffic flow prediction. Since we do not have auxiliary data, we compare the S-Match of RegionTrans. For each region in target city, RegionTrans uses short period data to calculate the linear similarity value with each region in source city. Then, the similarity is used as regularization for fine-tuning. For fair comparison, the base model of RegionTrans (i.e., the ST-net) is same as MetaST. 
    \item \textbf{MAML~\cite{finn2017model}}: an state-of-the-art meta-learning method, which learns a better initialization from multiple tasks to supervise the target task. MAML uses the same base model (i.e., the ST-net) as MetaST.
\end{itemize}

\subsubsection{Experimental Settings}~\\
\label{sec:parameter_traffic}
\textbf{Hyperparameter Setting.} 
For ST-net, we set the number of filters in CNN as 64, the size of map in CNN as $7\times 7$, the number of steps in LSTM as 8, and the dimension of hidden state of LSTM as 128. In the training process of taxi volume prediction, we set the learning rate of inner loop and outer loop as $10^{-5}$ and $10^{-5}$ respectively. For bike volume prediction, we set the learning rate of inner loop and outer loop as $10^{-5}$ and $10^{-6}$ respectively. The parameter $\gamma$ is set as $10^{-4}$. The number of updates for each task is set as 5. All the models are trained by Adam. The training batch size for each meta-iteration is set as 128, and the maximum of iteration of meta-learning is set as 20000. ~\\
\textbf{Spatial-temporal Clustering.} In addition, since the pattern of traffic volume usually repeats every day. Thus, in this work, we use averaged 24-hour patterns of each region to represent its spatial-temporal pattern. We use K-means to cluster these patterns to 4 groups. Furthermore, in this work, we do not use other external features, which means that $[\mathbf{\hat{y}}_{r_c, k_c}; \mathbf{e}_{r_c, k_c}]$ in Eq.~\eqref{eq:lstm} equals to $\mathbf{\hat{y}}_{r_c, k_c}$. We set the size of pattern representation in memory as 8.

\begin{table*}[!t]
\caption{Comparing with baselines for taxi volum prediction}
\centering
\begin{tabular}{|c|c|c|c|c|c|c|c||c|c|c|}
\hline
\multicolumn{2}{|c|}{\multirow{3}{*}{Taxi Data}} & \multicolumn{6}{c||}{Chicago}                                 & \multicolumn{3}{c|}{Boston}  \\ \cline{3-11} 
\multicolumn{2}{|c|}{}                           & \multicolumn{3}{c|}{Pick-up} & \multicolumn{3}{c||}{Drop-off} & \multicolumn{3}{c|}{Pick-up} \\ \cline{3-11} 
\multicolumn{2}{|c|}{}                           & 1-day    & 3-day   & 7-day   & 1-day    & 3-day    & 7-day   & 1-day    & 3-day   & 7-day   \\ \hline
\multicolumn{2}{|c|}{HA}      &    2.83   &   2.36  &  2.18   &  2.67   &  2.28    &  2.13  &   11.07 &  9.13 &  7.71 \\ \hline
\multicolumn{2}{|c|}{ARIMA}       & 3.19     & 2.76    & 2.71    & 2.93    & 2.43     & 2.41    & 11.02    & 10.25   & 9.36   \\ \hline
\multicolumn{2}{|c|}{ST-net}                     & 11.14    & 6.40    & 4.12    & 11.89    & 6.81     & 4.22    & 30.01    & 17.02   & 13.28   \\ \hline\hline
\multirow{3}{*}{Single-FT}        & NYC          & 2.88     & 2.18    & 1.86    & 3.01     & 2.91     & 2.00    & 11.71    & 10.62     & 7.76    \\ \cline{2-11} 
                                  & DC           & 4.13      & 2.78    & 2.17    & 4.42     & 2.32     & 2.28    & 14.39    & 12.32    & 9.19   \\ \cline{2-11} 
                                  & Porto        & 2.72    & 1.98   & 1.70     & 3.04     & 2.15     & 1.84    & 14.12    & 11.56    & 7.63    \\ \hline
\multicolumn{2}{|c|}{Multi-FT}                   & 2.31     & 2.01    & 1.69     & 2.33      & 2.20     & 1.79    & 9.90      & 9.41    & 6.68    \\ \hline
\multirow{3}{*}{RegionTrans}        & NYC          &  2.68    &  2.13   &   1.79  &  2.99   &    2.71 &   1.82  &   11.12  &  10.49    &  7.49   \\ \cline{2-11} 
 & DC  &   4.10   &   2.66  &  2.16  &   4.18   &   2.29  & 2.15   &   13.98  &  11.83   &  8.87  \\ \cline{2-11} 
 & Porto    &   2.60  &  1.94  & 1.70   &  3.02   &   2.09  &  1.83  &  13.05  &   10.97  &  7.43   \\ \hline
\multicolumn{2}{|c|}{MAML}                       & 2.13     & 1.89    & 1.61    & 2.23      & 2.04    & 1.76    & 8.13     & 7.59     & 5.88    \\ \hline\hline
\multicolumn{2}{|c|}{MetaST}                     &         \textbf{2.06}**       &   \textbf{1.80}**    &   \textbf{1.57}**    &      \textbf{2.16}**              &    \textbf{1.90}**   &   \textbf{1.75}*    &         \textbf{7.48}**        &        \textbf{7.15}**           &    \textbf{5.67}**    \\ \hline

\end{tabular}
\\\vspace{0.15cm}
	 ** (*) means the result is significant according to Student's T-test at level 0.01 (0.05) compared to MAML
\label{tab:restaxi}
\end{table*}

\begin{table}[!t]
\caption{Comparing with baselines for Bike Volume Prediction.}
	\centering
\begin{tabular}{|c|c|c|c|c|c|c|c|}
\hline
\multicolumn{2}{|c|}{\multirow{3}{*}{Bike Data}} & \multicolumn{6}{c|}{Chicago}          \\ \cline{3-8} 
\multicolumn{2}{|c|}{}                           & \multicolumn{3}{c|}{Pick-up}                                                                     & \multicolumn{3}{c|}{Drop-off}        \\ \cline{3-8} 
\multicolumn{2}{|c|}{}                           & 1-day                          & 3-day                          & 7-day                          & 1-day                          & 3-day                          & 7-day    \\ \hline
\multicolumn{2}{|c|}{HA}   &  4.97   &  3.69   &   3.64  &   4.96  &   3.67   & 3.63 \\ \hline
\multicolumn{2}{|c|}{ARIMA}                      & 4.86                           & 4.89                           & 4.79                           & 4.83                           & 4.97                           & 4.86     \\ \hline
\multicolumn{2}{|c|}{ST-net}                     & 7.61                           & 5.57                           & 3.83                           & 8.03                           & 5.45                           & 3.51  \\ \hline\hline
\multirow{2}{*}{SFT$^1$}         & NYC         &    2.52          &         2.49     &       1.95      &     2.49       &         2.41     &        1.87     \\\cline{2-8} 
                                   & DC          & 1.88                           & 1.99                           & 1.69                           & 2.03                           & 2.20                           & 1.67    \\\hline
\multicolumn{2}{|c|}{Multi-FT}                   & 1.97                           & 2.05                           & 1.62                           & 1.90                           & 1.98                           & 1.59    \\ \hline
\multirow{2}{*}{RT$^1$}         & NYC         &    2.50          &         2.23     &       1.87      &     2.36       &         2.18     &        1.76     \\\cline{2-8} 
& DC          &         1.86     &    1.95       &     1.66     &     1.98      &        2.09             &              1.63     \\ \hline
\multicolumn{2}{|c|}{MAML}                       & 1.85                           & 1.87                           & 1.62                           & 1.85                           & 1.79                           & 1.56         \\ \hline\hline
\multicolumn{2}{|c|}{MetaST}                      &          \textbf{1.76}**       &   \textbf{1.73}**    &   \textbf{1.45}**    &                    \textbf{1.83}**     &    \textbf{1.71}**   & \textbf{1.46}**  \\\hline
\end{tabular}
\\\vspace{0.15cm}
$^1$Due to the space limitation, in this table, SFT, RT mean single-FT and RegionTrans respectively.
\\\vspace{0.15cm}
** (*) means the result is significant according to Student's T-test at level 0.01 (0.05) compared to MAML.
\label{tab:resbike}
\end{table}

\subsubsection{Results}
We implement our model and compare it with the baselines on taxi and bike-sharing datasets. We run 20 testing times and report the average values. The results are shown in Table~\ref{tab:restaxi} and Table~\ref{tab:resbike}, respectively. According to these tables, we draw the following conclusions.
\begin{itemize}[leftmargin=*]\setlength{\itemsep}{0pt}
    \item Comparing with ST-net and some single-FT models, in some cases (e.g., 1-day training data), traditional time-series prediction methods (i.e., HA and ARIMA) achieves competitive performance in this problem. The reason is that traffic data show a strong daily periodicity, so that if we only have limited traffic data, we can still use periodicity to predict traffic volume.
    \item Comparing with ST-net, all transfer learning models (i.e., fine-tune models (including Single-FT and Multi-FT models), MAML, RegionTrans, MetaST) significantly improves the performance (i.e., lower the RMSE values). The results indicate that (1) it is difficult to train a model from random initialization with limited data; (2) the knowledge transfer between cities is effective for prediction. 
    \item In most cases, Multi-FT outperforms Single-FT. One possible reason is that Multi-FT increases the diversity of source domain. In other cases (e.g., Chicago pick-up prediction with 3-day training data), the best performance of Single-FT outperforms Multi-FT. The results implicitly indicates that if there exists a source city that is optimally transferable to the target city, simply mixing other cities may hurt the performance.
    \item RegionTrans models only slightly outperform their corresponding fine-tuning models (i.e., RegionTrans from NYC to Chicago v.s., Single-FT from NYC to Chicago). The results suggest that regional linear similarity regularization may not capture complex relationship between regions. In addition, since the data from different cities are collected from different time, regional similarity calculations may be inaccurate. Thus, it is not an effective and flexible way to transfer knowledge via regional similarity regularization.
    \item MAML and MetaST achieve better performance than fine-tuning methods and RegionTrans. This is because fine-tuning methods and RegionTrans cannot be adapted to different scenarios, and thereby decreasing the stability of transfer. However, MAML and MetaST not only learn the initialization based on multiple cities, but also achieve the best performance in every scenario sampled from these cities.
    \item Our proposed MetaST achieves the best performance in all experimental settings. Comparing with MAML, the averaged relative improvement is 5.0\%. 
    This is because our model can also capture and transfer long-term information, besides learning a better initialization. Moreover, the long-term pattern memory helps learn a further enhanced initialization. The stability of knowledge transfer increases to the highest degree. 
    \item Finally, we compare the results under different training data size in target city (i.e., 1-day, 3-day, and 7-day data). For every learning models (except HA and ARIMA), the performance improves with more training data. Our proposed MetaST still outperforms all baselines.
\end{itemize}

\subsubsection{Parameter Sensitivity}
MetaST involves a number of hyper-parameters. In this subsection, we evaluate how different selections of hyper-parameters impact our model's performance. Specifically, we analyze the impacts of two key parameters of spatial-temporal memory, i.e., the dimension $d$ of pattern representation and the trade-off factor $\gamma$ of two losses in the joint objective. All other hyperparameters are set as introduced in Section~\ref{sec:parameter_traffic}. We use the scenario of 3-day data for sensitivity analysis. 
\begin{figure}[t]
	\centering
	\begin{subfigure}[b]{0.23\textwidth}
		\centering
		\includegraphics[height=0.8\textwidth]{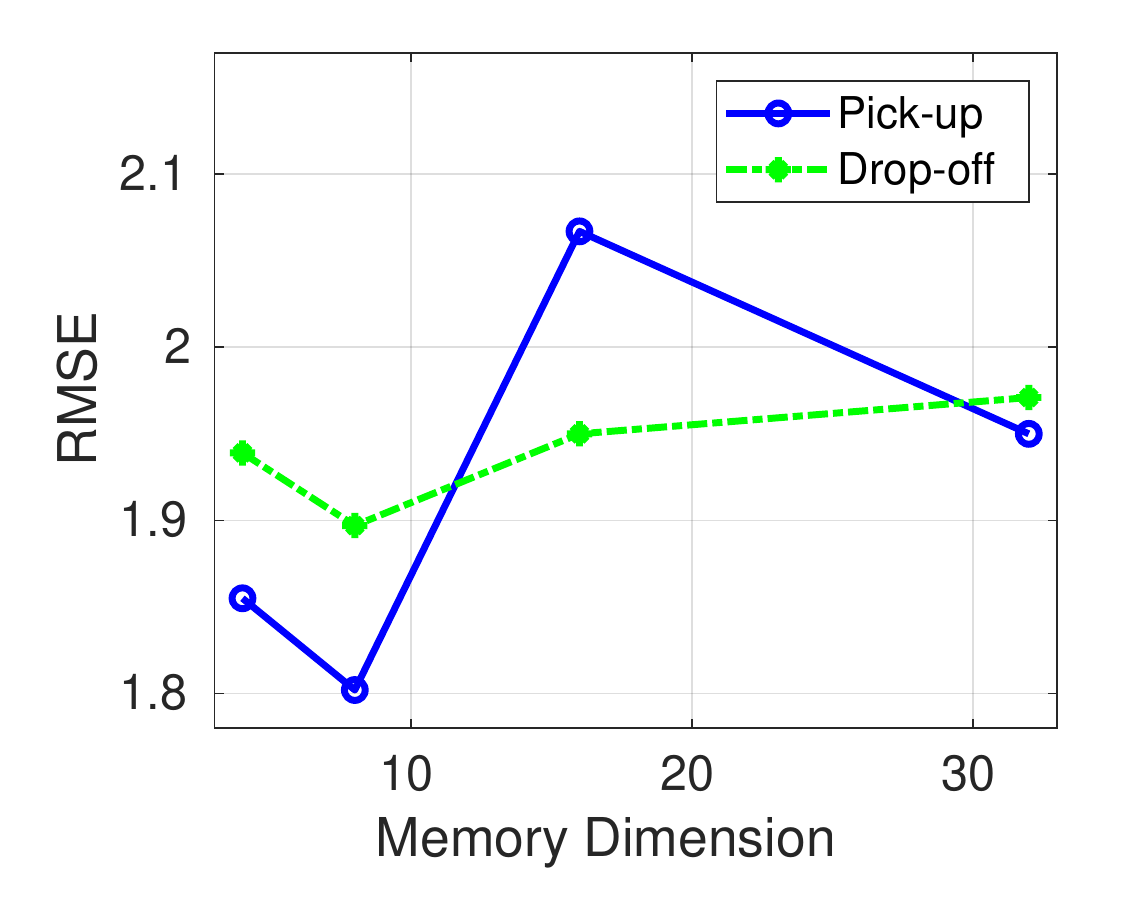}
		\caption{\label{fig:changedim}}
	\end{subfigure}
	\begin{subfigure}[b]{0.23\textwidth}
		\centering
		\includegraphics[height=0.8\textwidth]{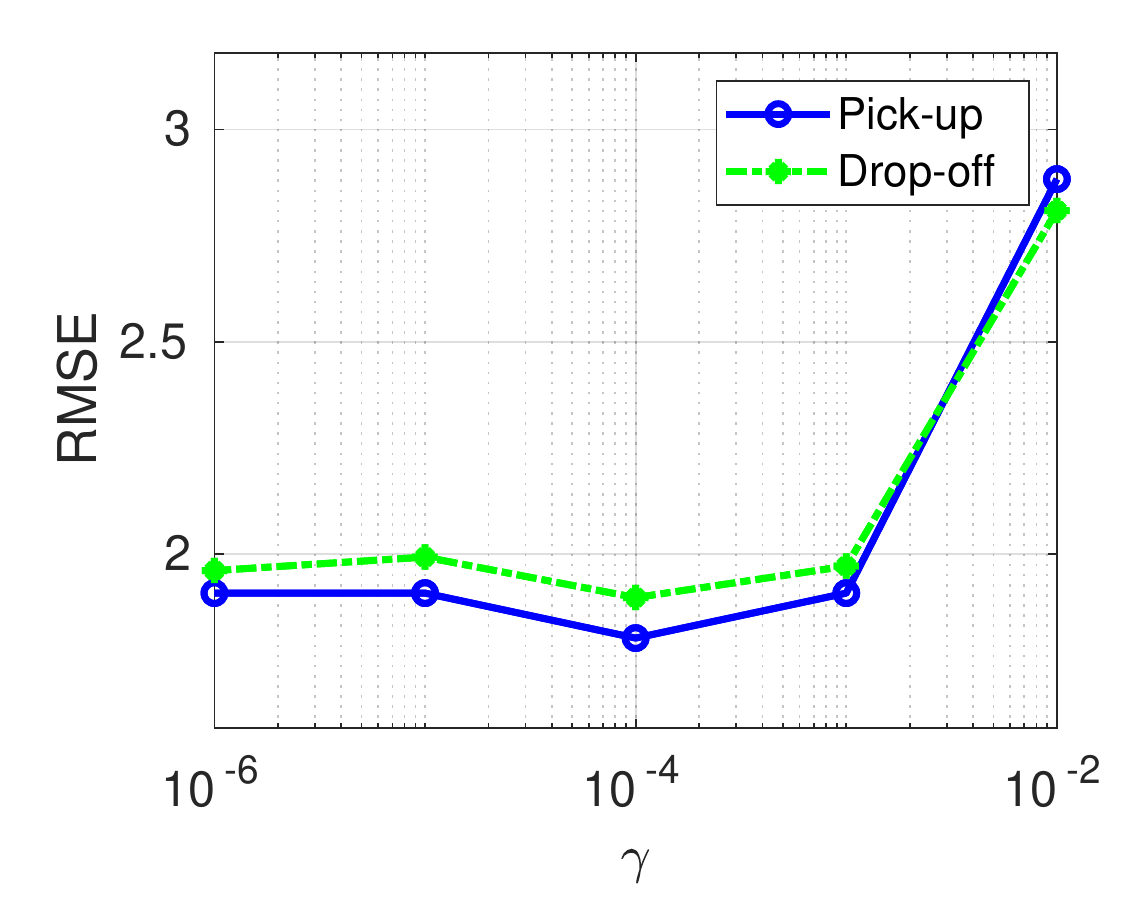}
		\caption{\label{fig:changegamma}}
	\end{subfigure}
	\begin{subfigure}[b]{0.23\textwidth}
		\centering
		\includegraphics[height=0.8\textwidth]{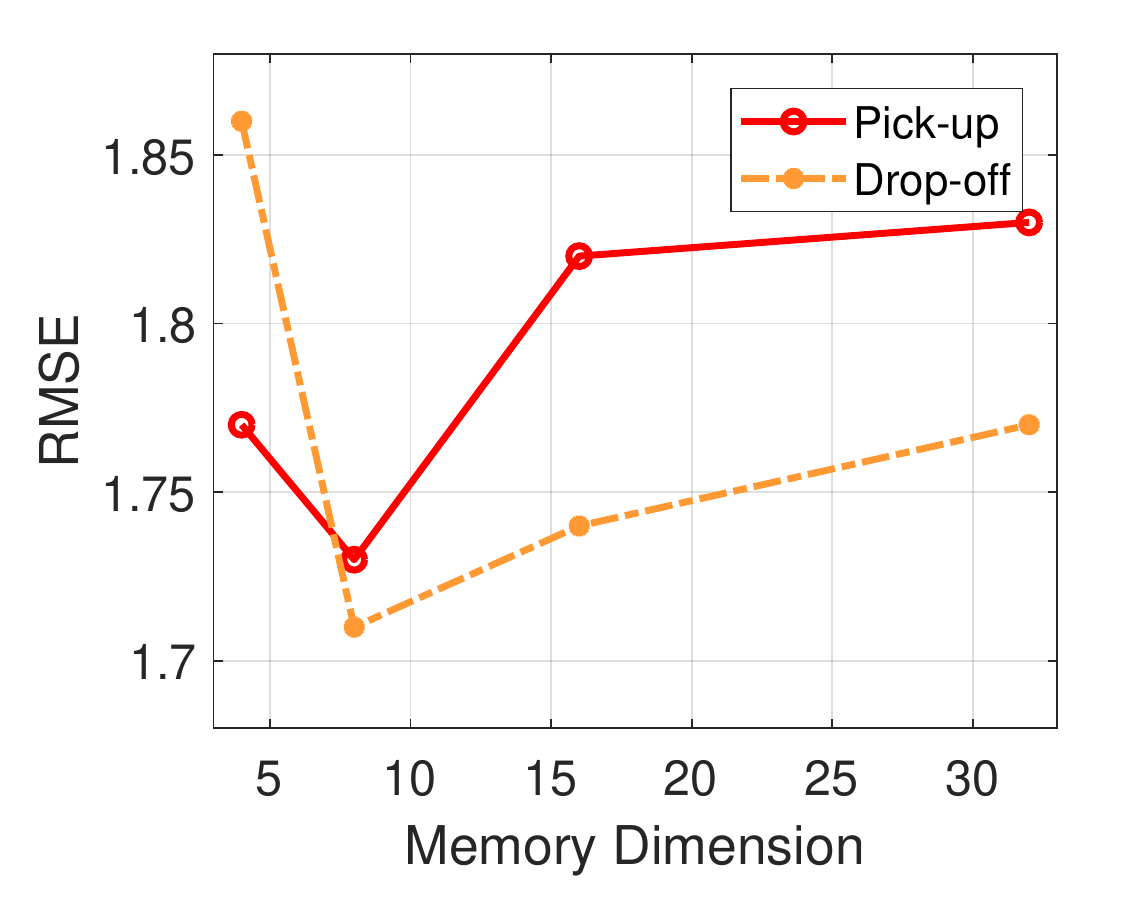}
		\caption{\label{fig:changedim_bike}}
	\end{subfigure}
	\begin{subfigure}[b]{0.23\textwidth}
		\centering
		\includegraphics[height=0.8\textwidth]{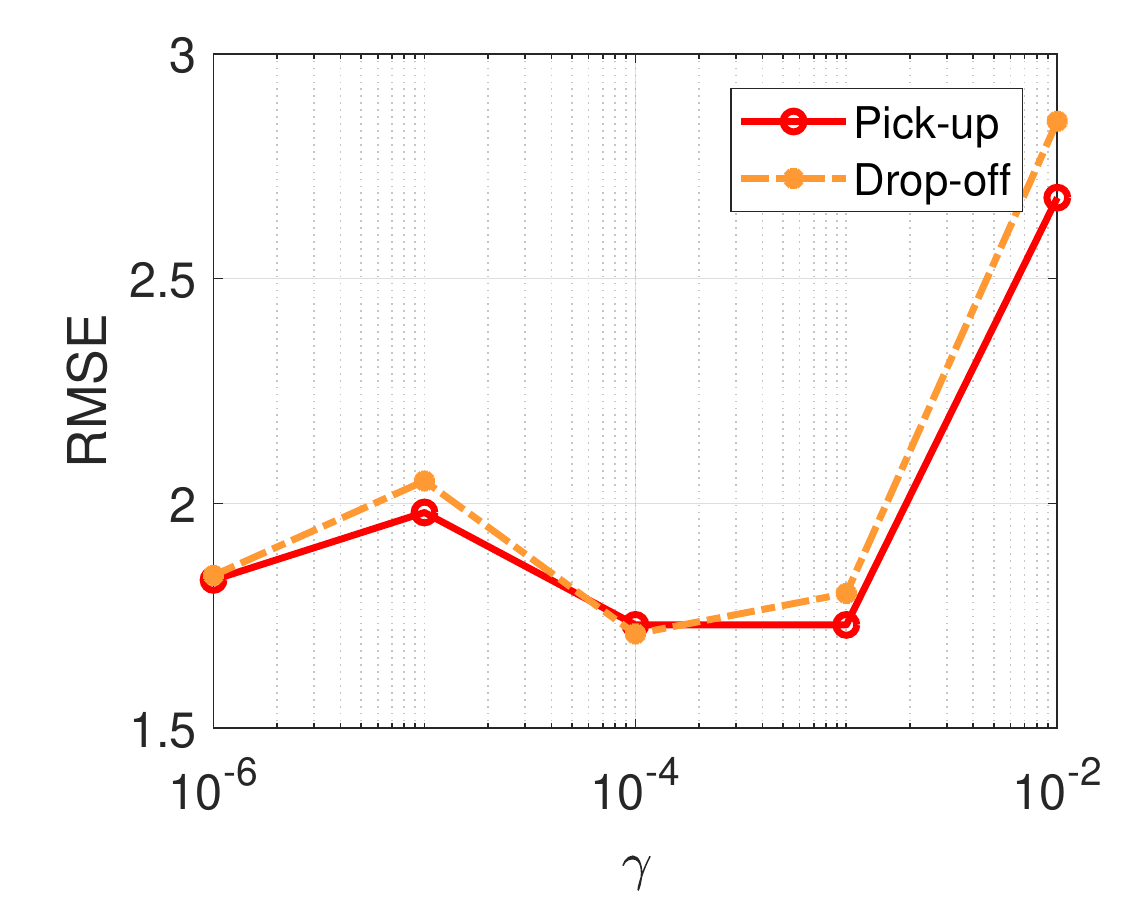}
		\caption{\label{fig:changegamma_bike}}
	\end{subfigure}
	\caption{(a) (c) RMSE with respect to the dimension of spatial-temporal memory in Chicago taxi/bike volume prediction, respectively; (b) (d) RMSE with respect to the value of $\gamma$ in Chicago taxi/bike volume prediction, respectively.} 
\end{figure}

For the the dimension $d$ of pattern representation, we change the dimension of pattern representation from 4 to 32 in spatial-temporal memory. The performance of Chicago taxi and bike volume prediction are shown in Figure~\ref{fig:changedim} and Figure~\ref{fig:changedim_bike}, respectively. We find that the performance increases in the beginning but decreases later. One potential reason is that the spatial-temporal memory provides too little information when the dimension is too small, while it can include too much irrelevant information when the dimension is too large. Both of the scenarios hurt the performance. Another experiment of trade-off factor $\gamma$ also demonstrates similar assumption. We change the parameter $\gamma$ in Eq.~\eqref{eq:loss} from $10^{-6}$ to $10^{-2}$. Higher value of $\gamma$ means higher importance of spatial-temporal memory. The results of Chicago taxi and bike volume prediction are shown in Figure~\ref{fig:changegamma} and Figure~\ref{fig:changegamma_bike}, respectively. We can see the similar change of the performance, increasing at first but decreasing later.
\begin{figure}[h]
	\centering
    \begin{subfigure}[b]{0.40\textwidth}
		\centering
		\includegraphics[height=0.6\textwidth]{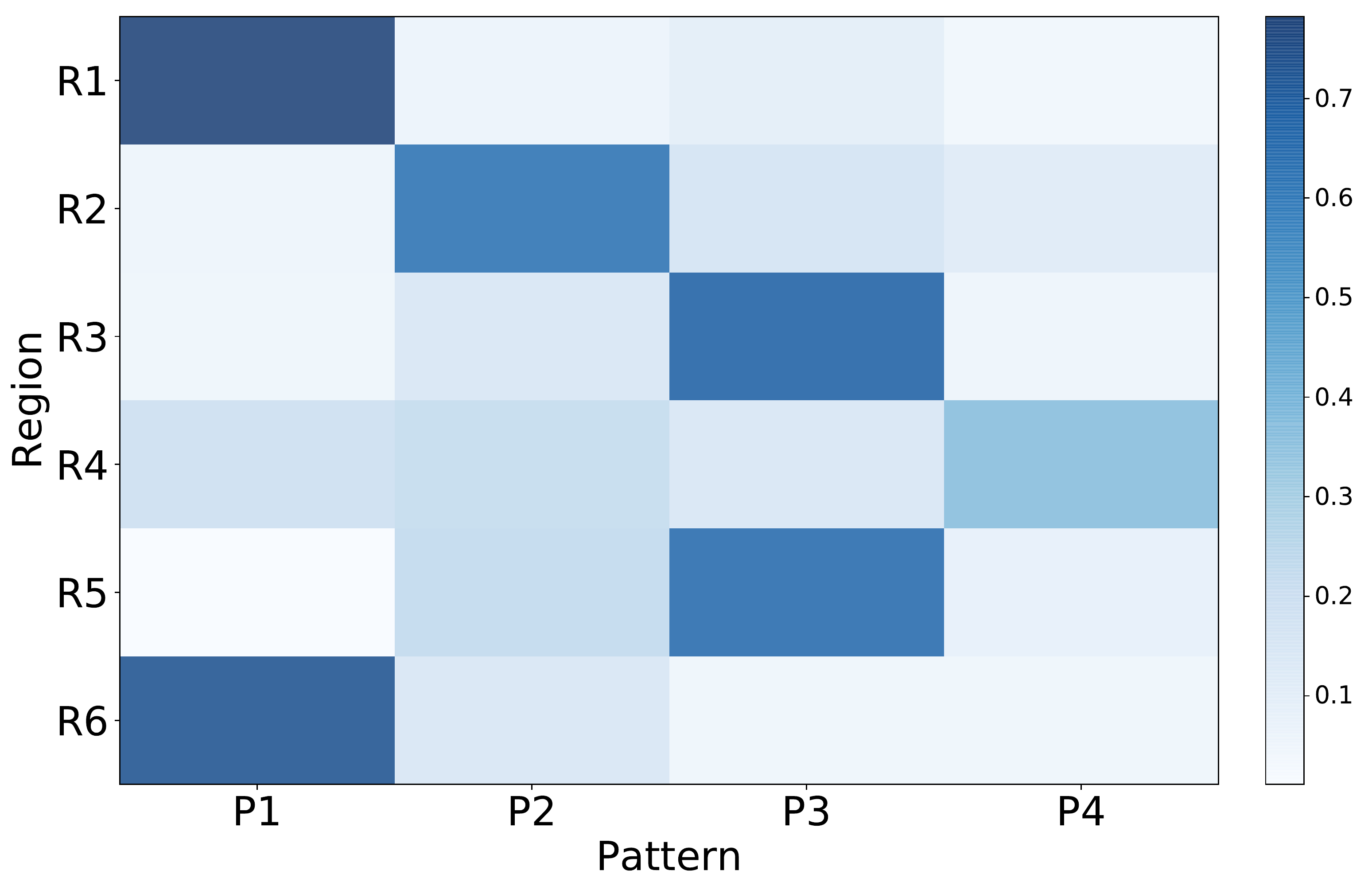}
		\caption{\label{fig:attvalue}}
	\end{subfigure}
	\begin{subfigure}[b]{0.15\textwidth}
		\centering
		\includegraphics[height=0.8\textwidth]{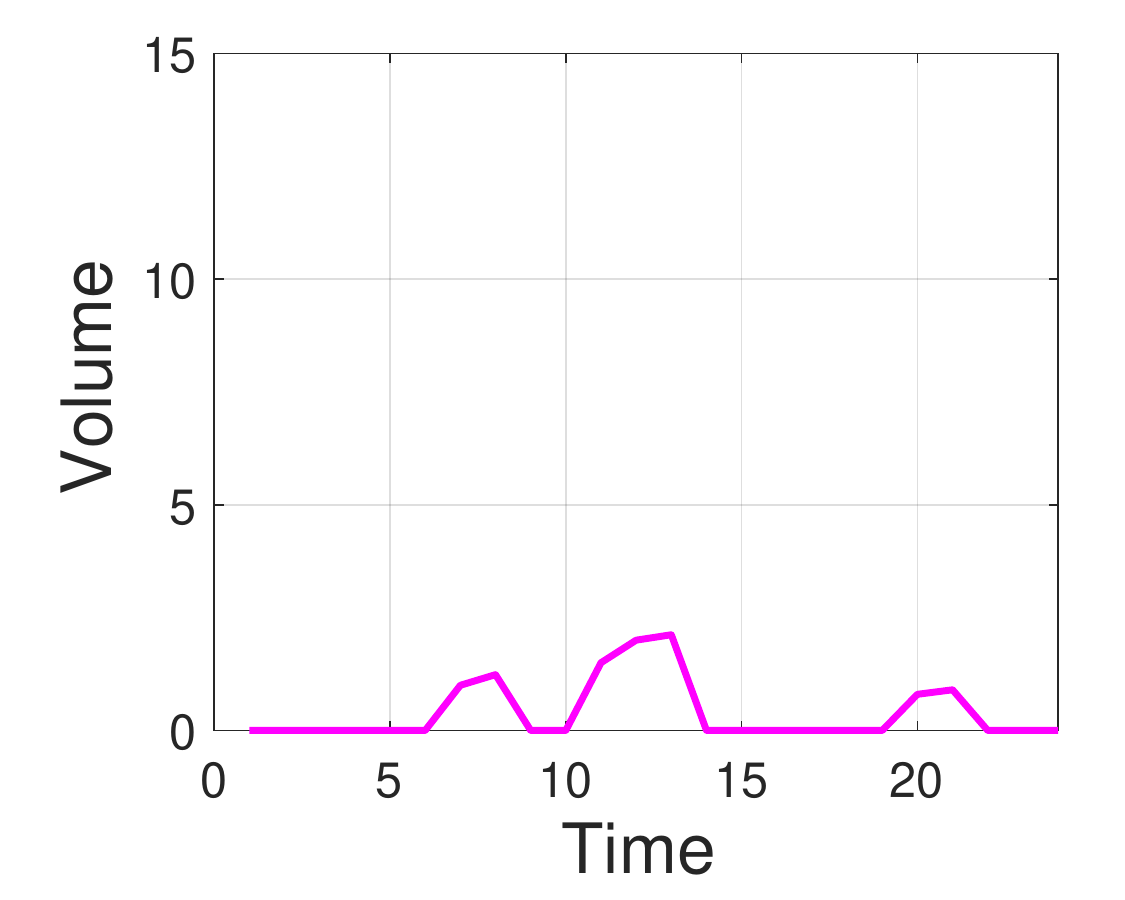}
		\caption{\label{fig:pattern0}}
	\end{subfigure}
	\begin{subfigure}[b]{0.15\textwidth}
		\centering
		\includegraphics[height=0.8\textwidth]{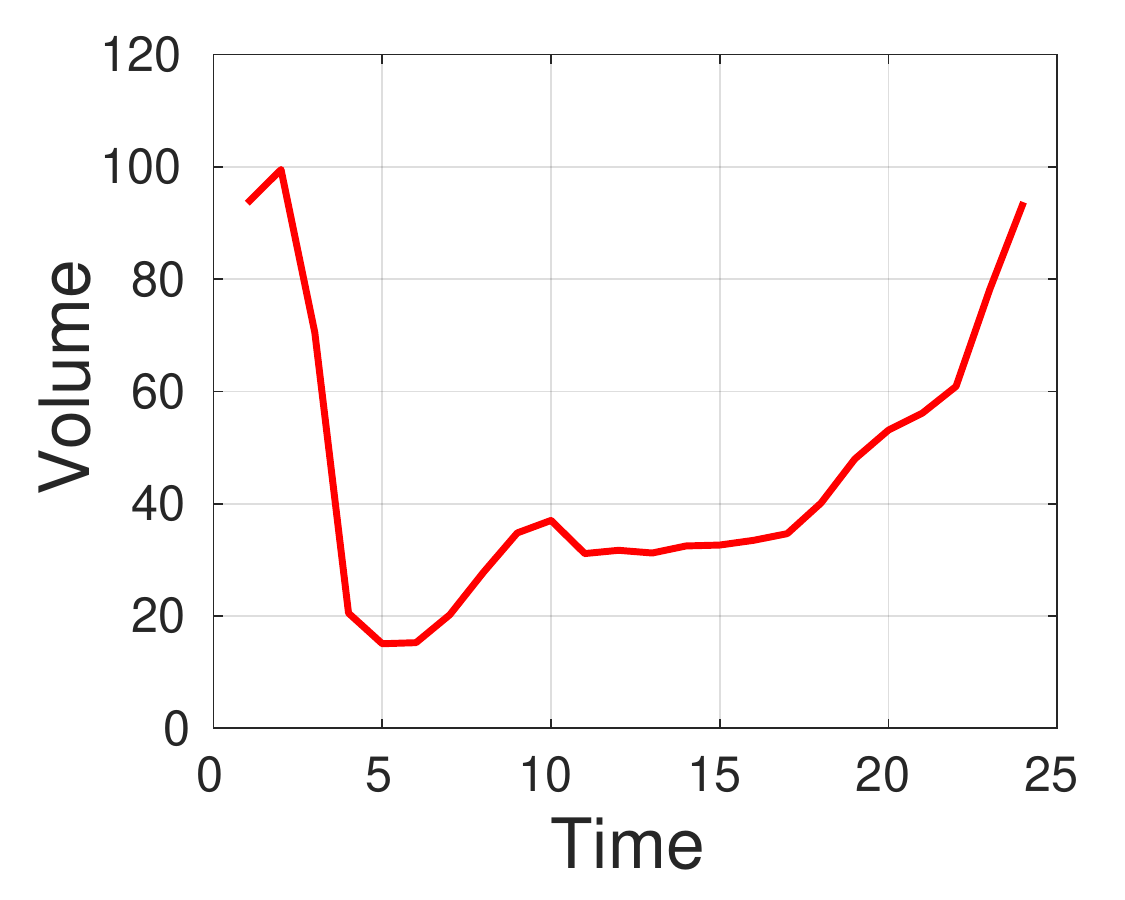}
		\caption{\label{fig:pattern1}}
	\end{subfigure}
    	\begin{subfigure}[b]{0.15\textwidth}
		\centering
		\includegraphics[height=0.8\textwidth]{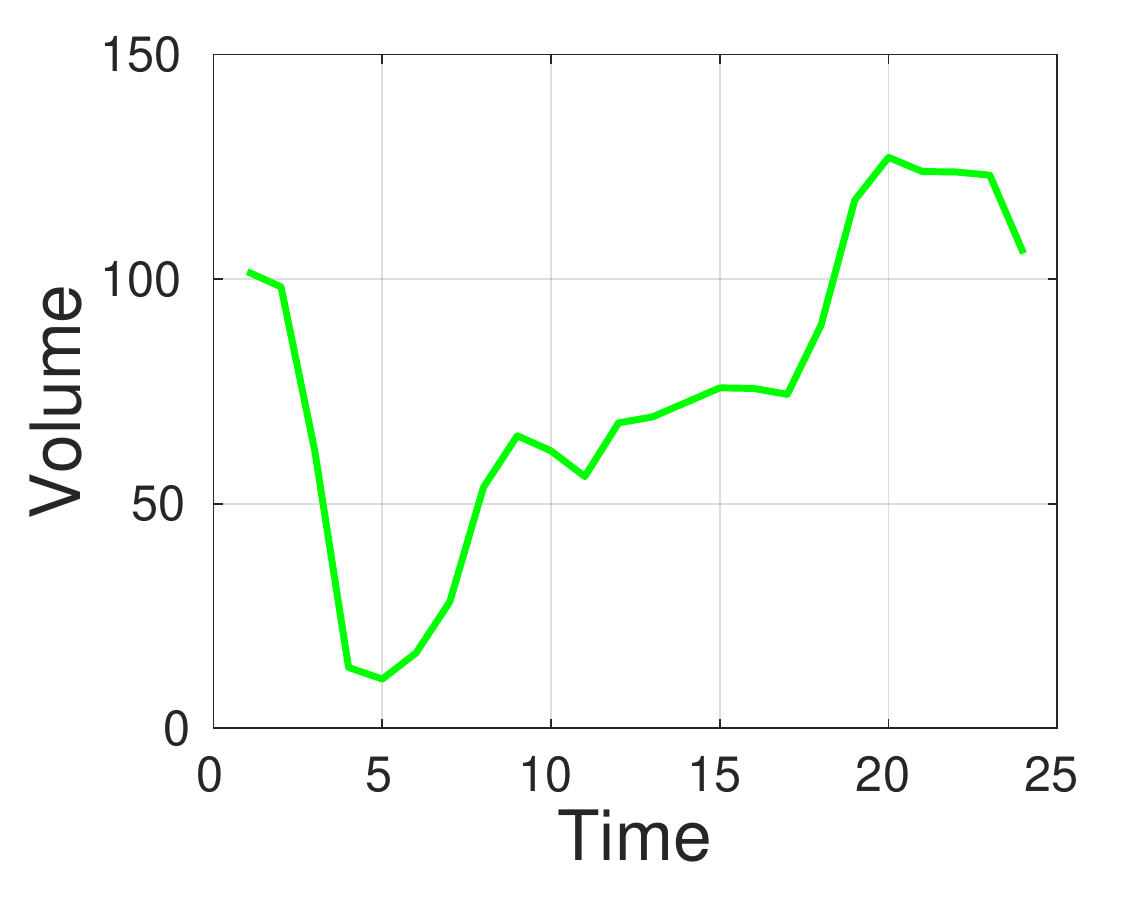}
		\caption{\label{fig:pattern2}}
	\end{subfigure}
	\begin{subfigure}[b]{0.15\textwidth}
		\centering
		\includegraphics[height=0.8\textwidth]{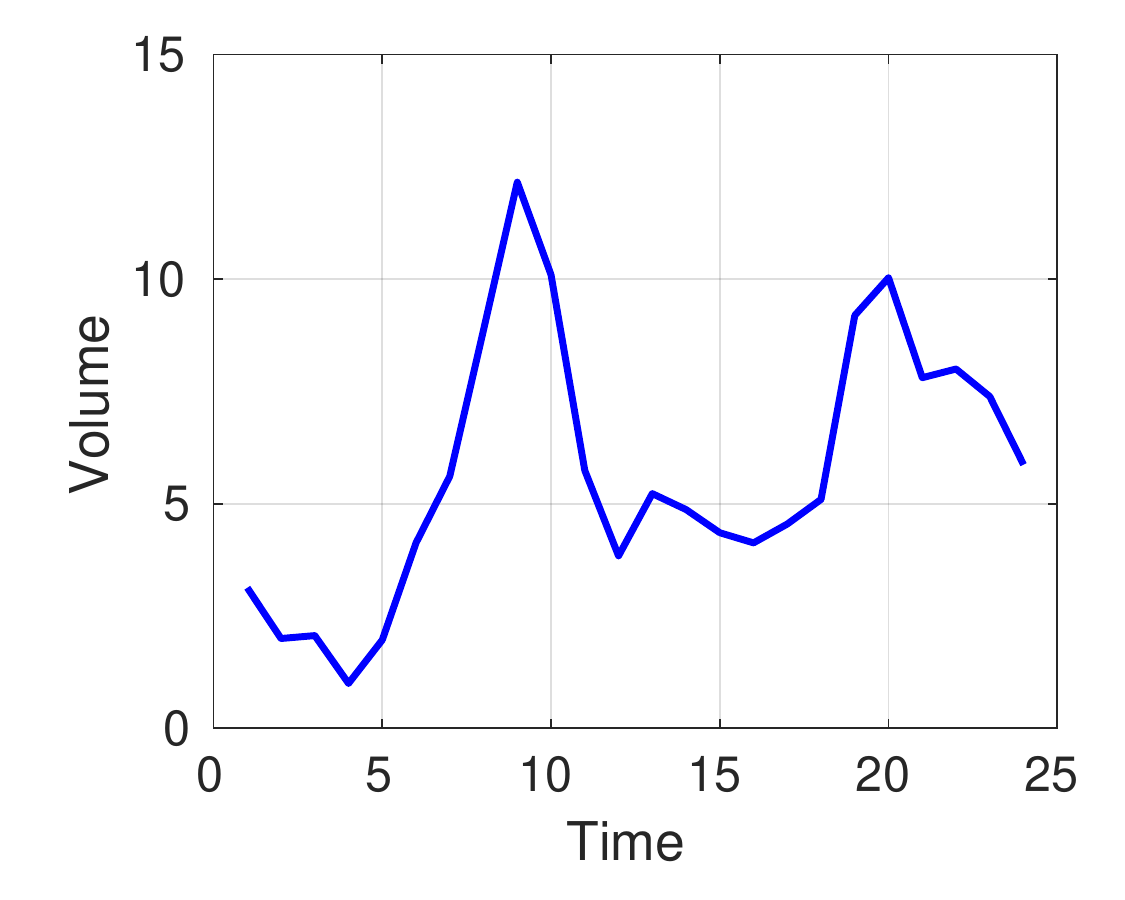}
		\caption{\label{fig:pattern3}}
	\end{subfigure}
	\begin{subfigure}[b]{0.15\textwidth}
		\centering
		\includegraphics[height=0.8\textwidth]{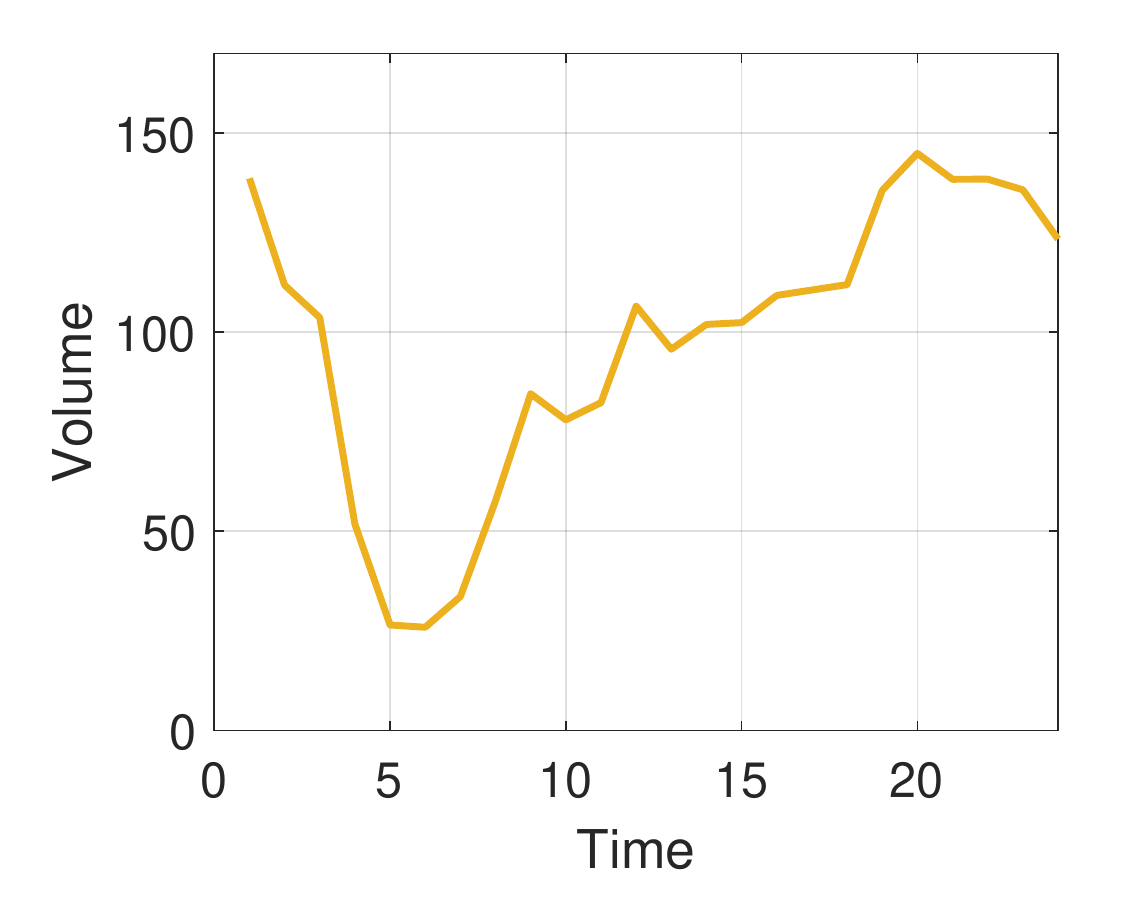}
		\caption{\label{fig:pattern4}}
	\end{subfigure}
	\begin{subfigure}[b]{0.15\textwidth}
		\centering
		\includegraphics[height=0.8\textwidth]{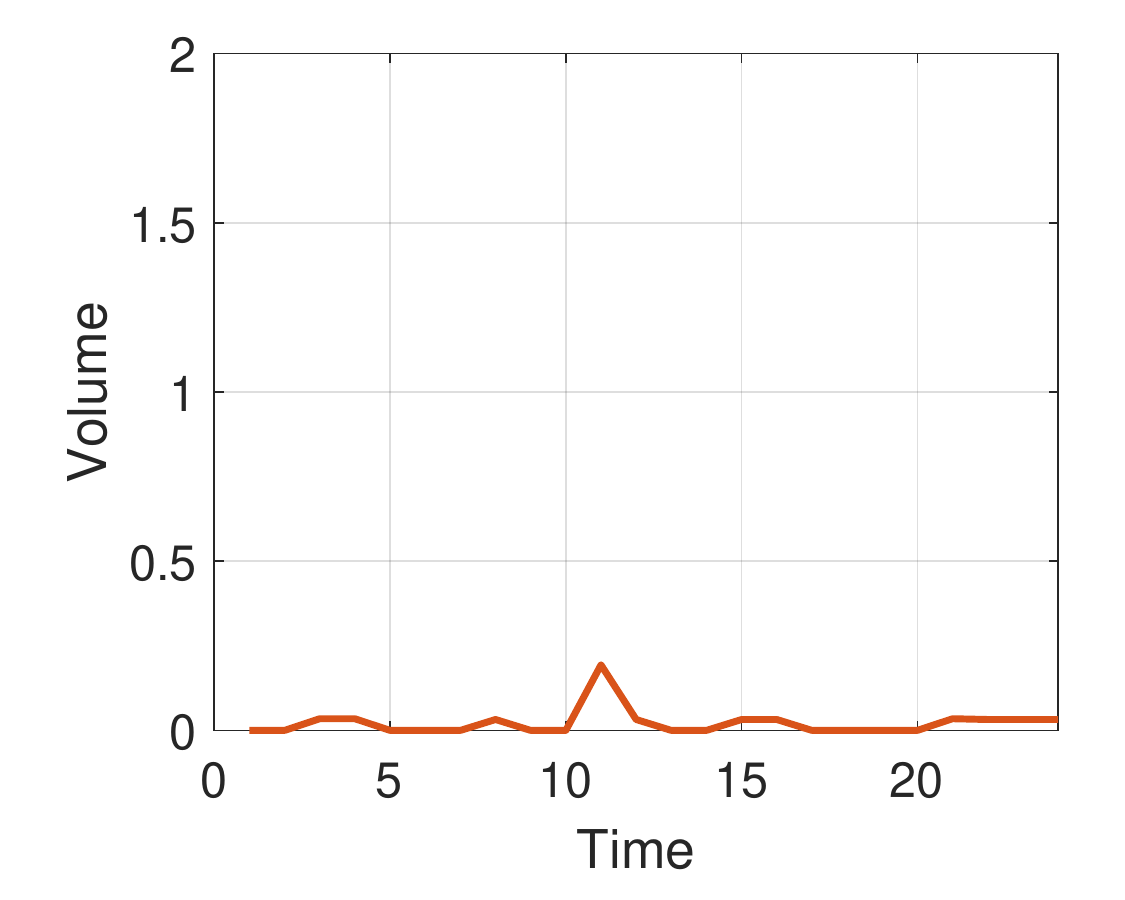}
		\caption{\label{fig:pattern5}}
	\end{subfigure}
	\caption{Spatial-temporal patterns detected by the memory $\mathcal{M}$ in Boston Taxi data. (a): probability of 4 pattern values of 6 regions; (b),(c),(d),(e),(f),(g) denote actual patterns of R1, R2, R3, R4, R5, R6, respectively.} 
    \label{fig:pattern}
\end{figure}

\subsubsection{Case Study: Visualization of Detected Patterns}
To intuitively demonstrate the efficacy of the usage of the Spatial-temporal memory, we visualize patterns detected from Boston taxi pick-up volume prediction. We also use 3-day data for this case study. We randomly select six regions and show the similarity values with respect to each pattern category in the memory $\mathcal{M}$. The results are shown in Figure~\ref{fig:attvalue}. The corresponding actual patterns of each region are shown in Figure~\ref{fig:pattern0}, Figure~\ref{fig:pattern1}, Figure~\ref{fig:pattern2}, Figure~\ref{fig:pattern3}, Figure~\ref{fig:pattern4}, and Figure~\ref{fig:pattern5} respectively. We can see that the taxi pick-up volume of R1 and R6 is almost zero and their attention weights are also similar (Pattern 1 is activated). The volume distribution in R2, R3 and R5 have one peak. The peak time of R2 is around 1:00am - 2:00am (Pattern 2 is activated). The pattern and attention weights of R3 and R5 are similar and the peak time is around 8:00pm - 9:00pm (Pattern 3 is activated). In R4, the volume distribution has two peaks (Pattern 4 is activated). One peak is around 9:00am - 10:00am, another one is 8:00pm - 9:00pm. The results indicate that the memory can distinguish regions with different patterns.
\begin{table*}[t]
\caption{Comparing with baselines for water quality prediction}
\centering
\begin{tabular}{|c|c|c|c|c||c|c|c||c|c|c|}
\hline
\multicolumn{2}{|c|}{\multirow{2}{*}{pH Data}} & \multicolumn{3}{c||}{Northeast}                                                                      & \multicolumn{3}{c||}{Southwest}                                                                      & \multicolumn{3}{c|}{South}                                                                          \\ \cline{3-11} 
\multicolumn{2}{|c|}{}                         & 1-year                          & 3-year                          & 7-year                          & 1-year                          & 3-year                          & 7-year                          & 1-year                          & 3-year                          & 7-year                          \\ \hline
\multicolumn{2}{|c|}{HA}   &   2.302   &  2.112   &   2.016  &  3.051   &    2.811  &  2.770 &  2.402   & 2.141  & 2.028  \\ \hline
\multicolumn{2}{|c|}{ARIMA}                    & 2.309                          & 2.328                           &   2.212                         &        3.153                   & 3.178                           & 3.103                            & 2.372                           & 2.363                           & 2.209                           \\ \hline
\multicolumn{2}{|c|}{ST-net}                   & 4.536                           & 3.806                           & 1.850                           & 2.694                           & 2.094                           & 1.008                           & 4.237                           & 3.951                           & 1.662                           \\ \hline\hline
\multirow{3}{*}{Single-FT}         & West         & 1.236                           & 1.128                           & 0.862                           & 0.935                           & 0.716                           & 0.683                           & 1.138                           & 1.043                           & 0.837                           \\ \cline{2-11} 
                                   & Midwest        & 1.048                           & 1.004                           & 0.806                           & 0.791                           & 0.653                           & 0.622                           & 0.970                           & 0.951                           & 0.793                           \\ \cline{2-11} 
                                   & Pacific        & 1.249                           & 1.140                           & 0.854                           & 0.928                           & 0.711                           & 0.671                           & 1.172                           & 1.031                           & 0.835                           \\ \hline
\multicolumn{2}{|c|}{Multi-FT}                 & 1.010                           & 0.987                           & 0.798                           & 0.706                           & 0.587                           & 0.567                           & 0.909                           & 0.898                           & 0.730                           \\ \hline
\multirow{3}{*}{RegionTrans}         & West         &  1.233   &      1.115   &      0.853          &          0.924       &      0.698      &      0.682       &    1.121     &   0.993  &  0.826     \\ \cline{2-11} 
& Midwest        &          1.047      &      0.988              &         0.796      &         0.783      &     0.651    &     0.619    &      0.965        &     0.938   &  0.769  \\ \cline{2-11} 
& Pacific        &  1.243   &     1.098    &  0.851   &       0.916     &         0.693        &     0.652    &  1.128  &       1.012   &   0.813   \\ \hline
\multicolumn{2}{|c|}{MAML}                     & 0.997                           & 0.955                           & 0.784                           & 0.701                           & 0.579                           & 0.559                           & 0.907                           & 0.897                           & 0.710                           \\ \hline\hline
\multicolumn{2}{|c|}{MetaST}                  &   \textbf{0.903}**      &   \textbf{0.898}**     &    \textbf{0.758}**    &    \textbf{0.649}**      &     \textbf{0.541}**    &   \textbf{0.514}**      &    \textbf{0.820}**     &    \textbf{0.803}**     &    \textbf{0.650}**    \\ \hline
\end{tabular}
\\\vspace{0.1cm}
	 ** (*) means the result is significant according to Student's T-test at level 0.01 (0.05) compared to MAML
\label{tab:envres}
\end{table*}
\subsection{Application-$\uppercase\expandafter{\romannumeral2}$: Water Quality Prediction}
\subsubsection{Problem Overview} 
The second application studied in this work is water quality prediction task. We also conduct a water quality prediction experiment. In this scenario, the water quality is represented by pH value of water, because pH value is easier to measure than other chemical metrics of water quality. We aim at predicting pH value in a specific location of next month (i.e., the time interval of water quality prediction is one month), as the changing of pH indicates the relative changes in the chemical composition in water. RMSE is still used as the evaluation metric in this task.

\subsubsection{Dataset}
The data used in this experiment is collected from the Water Quality Portal\footnote{https://www.waterqualitydata.us/} of the National Water Quality Monitoring Council. It spans about 52 years from 1966 to 2017. Each record represents one surface water sample with longitude, latitude, date and pH value in a standard unit. The continental U.S. area is roughly split to six areas: Pacific, West, Midwest, Northeast, Southwest, South. Note that, in the water quality prediction task, the areas are treated as the cities in previous descriptions. 

In addition, due to the sparsity of sampling points, we split each area into a grid region map, the size of each grid being $0.5$\degree of latitude by $0.5$\degree of longitude. Thus, the map sizes of all the six areas are 25$\times$50, 30$\times$25, 35$\times$25, 30$\times$25, 50$\times$25, 45$\times$25, respectively. The pH value of each region is represented by the median value of all sampling points in this region. We select Pacific, West, Midwest as source areas, and Northeast, Southwest, South as target areas. 
\subsubsection{Baselines}
We use the same baselines as in the experiments for traffic prediction. Both non-transfer methods and transfer methods are used to evaluated our algorithm. Note that, when calculating HA, the relative time is monthly. For example, if we want to predict the water quality at May, HA is the average value of all pH values at May in training data.~\\

\subsubsection{Experimental Settings}~\\
\textbf{Hyperparameter Setting.} Similar as the traffic prediction application, we do have external features in the water quality prediction task. For the learning process of spatial-temporal framework, we set the maximum of iterations as 5000, the number of filters in CNN as 32, the dimension of hidden states of LSTM as 64 and the size of memory representation as 4. Other hyperparameters are the same as traffic prediction.~\\
\textbf{Spatial-temporal Clustering.} By analyzing the data, pH in the current month is strongly correlated with pH in the same month of previous year. Thus, we use the 12-month periodic pattern of each region. Similar as traffic prediction task, K-means is also used to cluster these patterns to 3 groups. DTW distance is used to measure the distance of K-means. 
\subsubsection{Results}
We implement our model and compare with baselines on the water quality prediction task. We run 20 testing tasks and report the average values in Table~\ref{tab:envres}.  
Most experiment results and their explanation are similar to traffic prediction. Besides, from this table, we observe that:
\begin{itemize}[leftmargin=0.15in]\setlength{\itemsep}{0pt}
    \item Comparing with Multi-FT model, the performance of MAML only slightly improves in most cases. The potential reason is that the regions in the source areas may be homogeneous in short-term performance. Thus, compared to MAML which learning a initialization, simply mixing all samples (i.e., Multi-FT) may not significantly hurt the performance significantly.
    \item MetaST achieves the best performance compared with all baselines with the averaged relative improvement as 7.7\%. Since MetaST provides more detailed long-term information by spatial-temporal memory and distills the long-term information to target city, which explicitly increases the diversity of regions in source domain. 
\end{itemize}

\subsubsection{Parameter Sensitivity}
\begin{figure}[t]
	\centering
	\begin{subfigure}[b]{0.235\textwidth}
		\centering
		\includegraphics[height=0.8\textwidth]{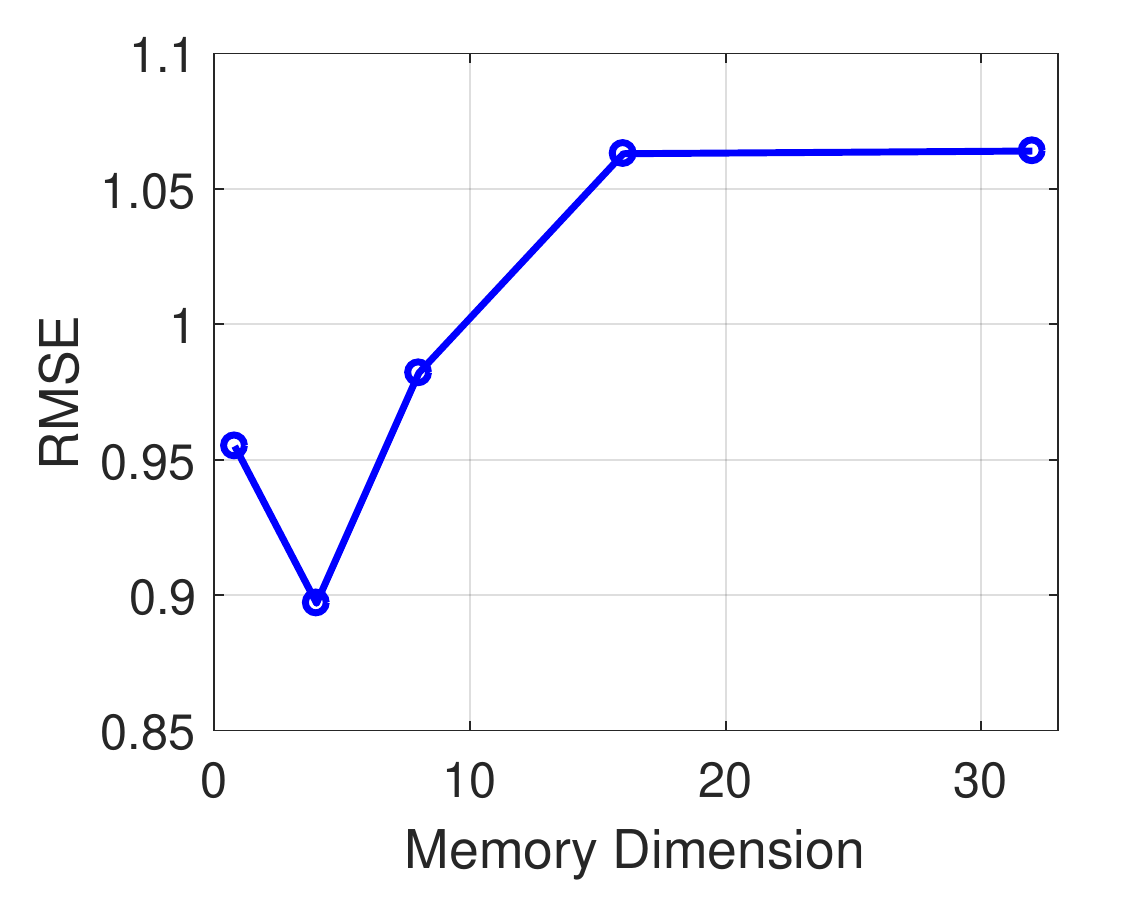}
		\caption{\label{fig:changedim_env}}
	\end{subfigure}
	\begin{subfigure}[b]{0.235\textwidth}
		\centering
		\includegraphics[height=0.8\textwidth]{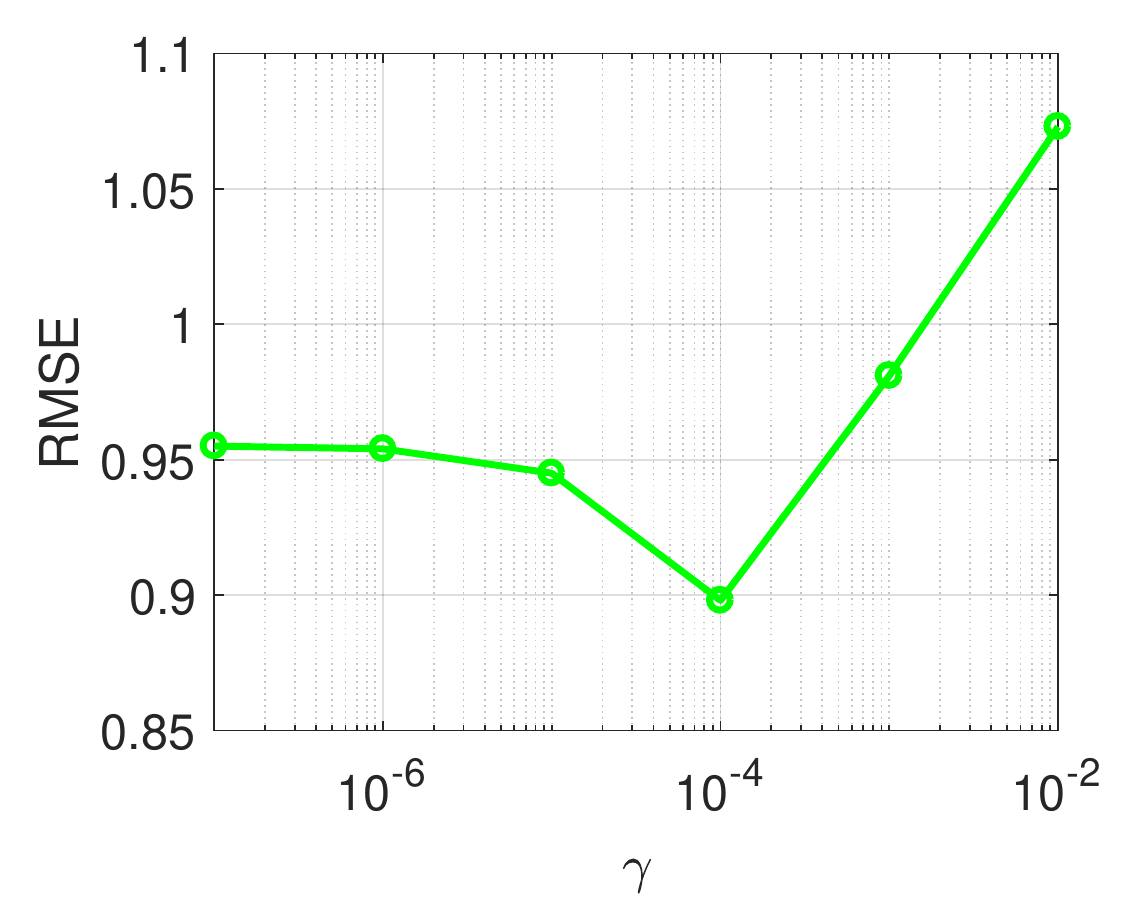}
		\caption{\label{fig:changegamma_env}}
	\end{subfigure}
	\caption{(a) RMSE with respect to the dimension of spatial-temporal memory; (b) RMSE with respect to the value of $\gamma$.} 
	\vspace{-1.5em}
\end{figure}

Following the same step of traffic prediction task, we investigate the effect of the dimension $d$ of pattern representation and the trade-off factor $\gamma$ of two losses in the joint objective on MetaST performance. The performance of $d$ from 2 to 32 and $\gamma$ from $10^{-7}$ to $10^{-2}$ on water quality value prediction is shown in Figure~\ref{fig:changedim_env} and Figure~\ref{fig:changegamma_env}, respectively. Both Figure~\ref{fig:changedim_env} and Figure~\ref{fig:changegamma_env} are evaluated on Northeast water quality prediction with 3-year training data. Accordingly, MetaST achieves the best performance when $d=4$ and $\gamma=10^{-4}$. Similar as the reasons in traffic prediction, the results indicate that suitable selection of $d$ and $\gamma$ lead to the best performance.

\section{Conclusion and Discussion}
In this paper, we study the problem of transferring knowledge from multiple cities for spatial-temporal prediction. We propose a novel MetaST model which leverages learned knowledge from multiple cities to help with the prediction in target data-scarce cities. Specifically, the proposed model learns a well-generalized initialization of spatial-temporal prediction model for easier adaptation. Then, MetaST  a global pattern-based spatial-temporal memory from all source cities.  We test our model on two spatial-temporal prediction problem from two different domains: traffic prediction and environment prediction. The results show the effectiveness and of our proposed model. 

For future work, we plan to investigate from two directions: (1) We plan to further consider network structure (e.g., road structure) and combine it with our proposed model. For example, a simple extension is that we can use graph convolutional network as our base model; (2) We plan to explain the black-box transfer learning framework, and analyze which information is transferred (e.g., spatial correlation, region functionality).

\begin{acks}
We thank Xiuwen Yi for feedback on an early draft of this paper. The work was supported in part by NSF awards \#1652525,
\#1618448, and \#1639150. The views and conclusions contained in
this paper are those of the authors and should not be interpreted
as representing any funding agencies.
\end{acks}
\bibliographystyle{ACM-Reference-Format}
\balance 
\bibliography{ref}
\end{document}